\documentclass[letterpaper]{article} 
\usepackage{aaai24}  
\usepackage{times}  
\usepackage{helvet}  
\usepackage{courier}  
\usepackage[hyphens]{url}  
\usepackage{graphicx} 
\usepackage{natbib}  
\usepackage{caption} 
\usepackage{algorithm}
\usepackage{algorithmic}

\usepackage{multicol}
\usepackage{multirow}
\usepackage{booktabs}
\usepackage{subcaption}

\usepackage{newfloat}
\usepackage{listings}
\DeclareCaptionStyle{ruled}{labelfont=normalfont,labelsep=colon,strut=off} 
\floatstyle{ruled}
\newfloat{listing}{tb}{lst}{}
\floatname{listing}{Listing}
\nocopyright

\title{TETRIS: Towards Exploring the Robustness of Interactive Segmentation}
\author {
    Andrey Moskalenko\thanks{Correspondence to and.v.moskalenko@gmail.com},
    Vlad Shakhuro,
    Anna Vorontsova,
    Anton Konushin,\\
    Anton Antonov,
    Alexander Krapukhin,
    Denis Shepelev,
    Konstantin Soshin
}
\affiliations {
    Samsung Research\\
    \{andrey.mos, v.shakhuro, a.vorontsova, a.konushin, ant.antonov, a.krapukhin, d.shepelev, k.soshin\}@samsung.com
}

\begin{document}

\maketitle

\begin{abstract}

Interactive segmentation methods rely on user inputs to iteratively update the selection mask. A click specifying the object of interest is arguably the most simple and intuitive interaction type, and thereby the most common choice for interactive segmentation. However, user clicking patterns in the interactive segmentation context remain unexplored. Accordingly, interactive segmentation evaluation strategies rely more on intuition and common sense rather than empirical studies (e.g., assuming that users tend to click in the center of the area with the largest error). In this work, we conduct a real user study to investigate real user clicking patterns. This study reveals that the intuitive assumption made in the common evaluation strategy may not hold. As a result, interactive segmentation models may show high scores in the standard benchmarks, but it does not imply that they would perform well in a real world scenario. To assess the applicability of interactive segmentation methods, we propose a novel evaluation strategy providing a more comprehensive analysis of a model's performance. To this end, we propose a methodology for finding extreme user inputs by a direct optimization in a white-box adversarial attack on the interactive segmentation model. Based on the performance with such adversarial user inputs, we assess the robustness of interactive segmentation models w.r.t click positions. Besides, we introduce a novel benchmark for measuring the robustness of interactive segmentation, and report the results of an extensive evaluation of dozens of models.

\end{abstract}

\begin{figure}[ht]
    \centering
    \includegraphics[width=1.0\columnwidth]{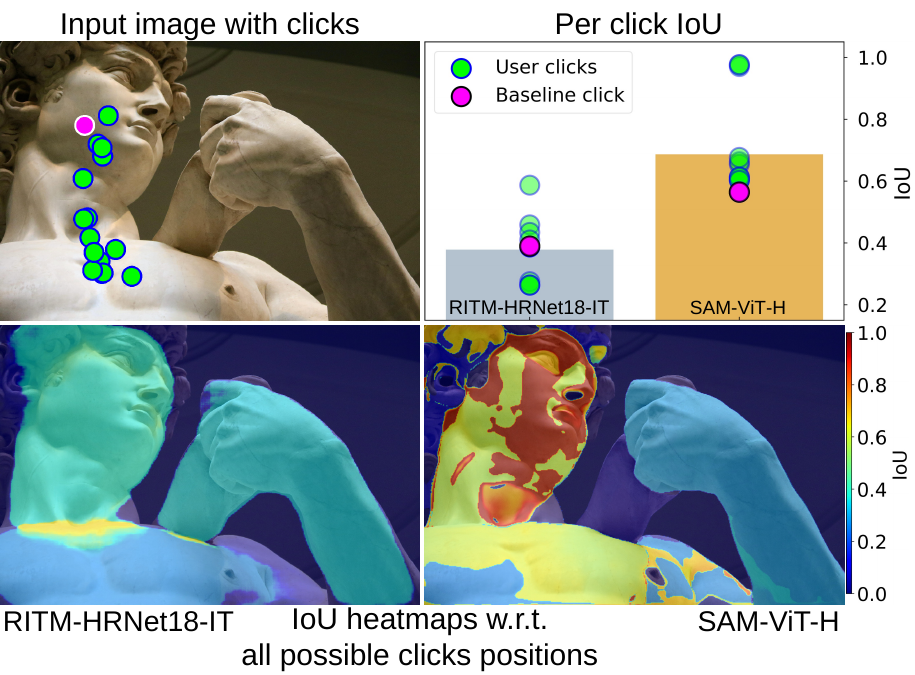}
    \caption{Single clicks made by different real users and the respective quality achieved. 
    Top left: real users (green dots) do not click the way it is assumed in the standard testing procedures (magenta dot). Top right: the quality of two popular interactive segmentation models, a convolutional RITM~\cite{ritm} and a transformer-based SAM~\cite{kirillov2023segment}, is widely spread around the average score (visualized with colored bars). Bottom: IoU heatmaps show that prediction quality fluctuates heavily depending on an actual click position.}
        \label{fig:clickbait}
\end{figure}

\begin{figure*}[ht]
    \centering
    \includegraphics[width=0.9885\textwidth]{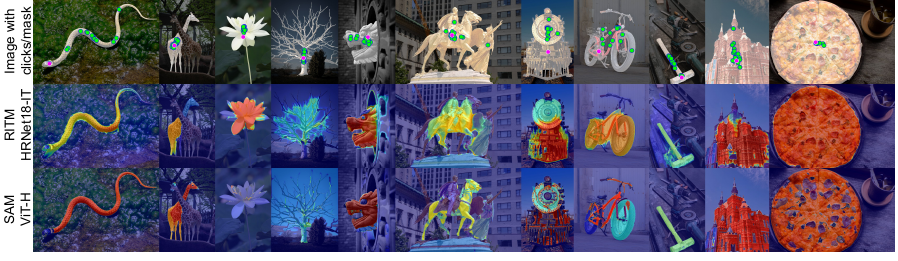}
    \caption{Top row: images with overlapped ground-truth masks (white), real user clicks (green), and clicks generated with the baseline strategy (magenta). Two bottom rows: IoU scores of RITM and SAM, calculated on a grid for each possible integer click position; warmer colors correspond to higher scores. Apparently, IoU scores may vary dramatically within small regions of the same object: this shows that the state-of-the-art approaches are rather sensitive to the click position.}
    \label{fig:user-study-visual}
\end{figure*}

\section{Introduction}

Interactive segmentation methods are widely exploited for object removal, object selection, large dataset collection, medical image annotation and other tasks related to image labeling. Compared to conventional segmentation approaches, interactive methods provide higher quality masks that satisfy user requests better. Arguably the most well-explored, click-based interactive segmentation aims at selecting objects in an image according to multiple user input clicks (either positive or negative), comprising a \textit{click trajectory}. However, the real user evaluation of each novel approach, that implies comparing it with an ever-growing number of predecessors, is completely unfeasible. Respectively, in the common interactive segmentation benchmarks, clicks are not put by real users but automatically generated based on a history of interactions. Most existing methods 1) select a region with the largest error in the previous interaction round, and 2) click in the furthest point from the boundaries of this region. Hereinafter, we refer to this click generation scheme as the \textit{baseline strategy}.

However, our study of real user clicks reveals this straightforward strategy does not emulate user behavior adequately. Besides, models tend to overfit to the baseline strategy, so that the accuracy might be high, but even a slight change of a click position causes a severe quality drop~(Figure~\ref{fig:clickbait}). Thus, the real-usage quality of the tested models remains unknown. 

The evaluation protocols that assess quality for a single click trajectory cannot guarantee that tested methods are robust enough to perform well in various possible interaction scenarios. In this work, we formulate a multi-trajectory evaluation strategy. Particularly, we propose to generate click trajectories through a differentiable adversarial attack on the interactive segmentation model, and estimate the robustness based on a quality gap between trajectories.

Overall, our contributions are as follows:

\begin{itemize}
    \item We conduct a pioneer study of real user clicking patterns in an interactive segmentation scenario. It reveals that users do not always click in the center of an area with the largest error, as assumed in the baseline methodology used in the most existing methods.
    \item To the best of our knowledge, we are the first to develop a procedure for generating user inputs via an adversarial attack for measuring robustness of interactive models. Relying on a differentiable rendering of user inputs, the proposed procedure remains fully differentiable and fast-convergent.
    \item We present a TETRIS benchmark with 2000 high-resolution images carefully selected and manually labeled with fine segmentation masks. The images depict common objects: 1000 images contain objects of various categories, and another 1000 portray people.
    \item We formulate an interactive segmentation robustness score, and evaluate the robustness of state-of-the-art methods, using TETRIS and the standard interactive segmentation benchmarks. 
\end{itemize}

We believe that the methodology presented in this study will assist creating more robust and high-quality interactive models for real world applications.

\begin{figure*}
    \centering
    \includegraphics[width=1.0\textwidth]{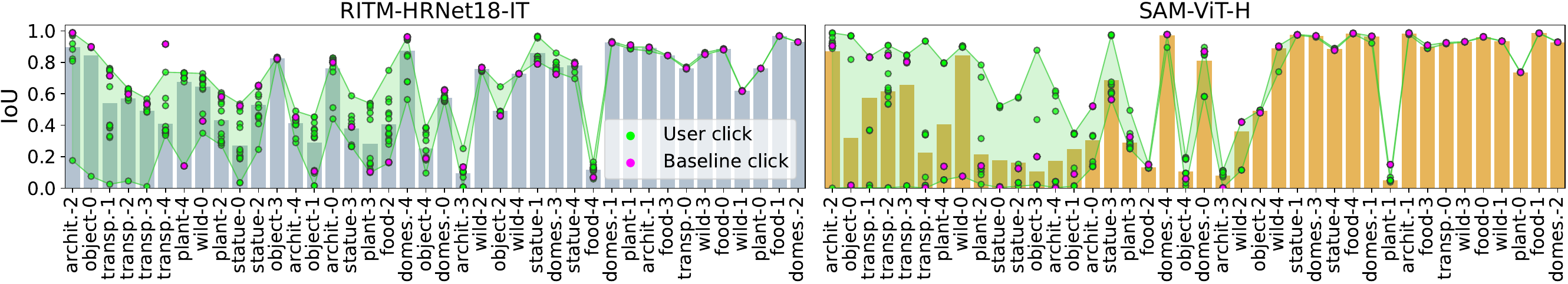}
    \caption{An IoU spread (a difference between a maximum and minimum IoU of user clicks) between predicted and ground truth masks in the first real user interaction round. Green points represent user clicks, magenta points depict the clicks generated with the baseline strategy. Columns are sorted by an average spread.}
    \label{fig:user-study-std}
\end{figure*}

\section{Related Work}

\paragraph{Benchmarking Interactive Segmentation.}
GrabCut \cite{grabcut} was the first interactive segmentation dataset. Then, the Berkeley \cite{berkeley} segmentation dataset was adapted for interactive segmentation~\cite{berkeley-intro}. The associated evaluation protocol implied assessing both object and boundary segmentation quality with IoU measure and required manual interaction with a method. \cite{deep-object-selection} proposed an automatic procedure of benchmarking click-based interactive segmentation on PASCAL VOC 2012 \cite{pascal-voc-2012} and COCO \cite{coco} segmentation datasets; in this procedure, clicks were placed strictly in the center of the largest erroneous region, and the quality was assessed with IoU. The follow-up work \citep{latent-diversity} adapted DAVIS \cite{davis} and SBD \cite{sbd} (labeled with boundaries) datasets for interactive segmentation, using the same click generation strategy. We conduct our study on most commonly used datasets as well as TETRIS, which contains images of a significantly higher resolution.

\paragraph{Segmentation Metrics.}

The most common metric used to assess interactive segmentation is the Number of Clicks (NoC) \cite{jang2019interactive, sofiiuk2020f}, required to achieve the predefined IoU score. NoC equally penalizes the cases where the desired score was achieved on the last interactions, and the cases where the threshold was not exceeded; we consider this to be a major drawback of this metric. Besides, it was noticed \cite{ritm} that using the baseline strategy encourages the model to overfit to NoC, while the performance in a real scenario remains poor.

Accordingly, we do not use NoC, but consider an area under an IoU curve \cite{jang2019interactive} as a major metric. We consider 10 clicks and normalize the area to be within $[0, 1]$. The standard IoU score is edge-insensitive, so boundary metrics were additionally formulated for an ad-hoc assessment. The trimap IoU \cite{kohli2009consistency, deeplab} is calculated within a distance from the ground truth mask boundary, ignoring distant erroneous pixels. The performance issue was addressed with approximations of F-measure \cite{csurka2013measure, perazzi2016video}. McGuinness et al. \cite{berkeley-intro} formulated a fuzzy boundary accuracy measure. \cite{cheng2020cascadepsp} proposed a mean Boundary Accuracy measure (mBA), further reworked into an MQ \cite{yang2020meticulous} score. In an image matting, boundary quality is evaluated with a trimap-based SAD, MSE \cite{xu2017deepmatting}, and perceptual Gradient and Connectivity errors \cite{xu2017deepmatting}. Since pronounced boundaries are especially important for high-resolution image processing, we also measure the boundary quality. To this end, we use a recently introduced Boundary IoU \cite{cheng2021boundary}, which is intuitive and one of the most straightforward. 

\paragraph{User Inputs.}

In this study, we consider only click-based approaches. However, numerous works were dedicated to other user input types. Bounding boxes were employed for selecting large image areas~\cite{deep-object-selection,grabcut}. In \cite{gueziri2017latency}, object selection was guided with manual strokes. In \cite{ferrari2019scribbles}, an initial selection was made using bounding boxes obtained via extreme clicking \cite{papadopoulos2017extreme}, and then refined with strokes. \cite{cheng2021flexible} proposed a randomized uniform click and stroke generation strategy, where points were randomly sampled from the ground truth mask. Recently presented Segment Anything, or SAM~\cite{kirillov2023segment}, formulated a \textit{promptable segmentation} task, where each prompt can be a point, a box, a mask, or a text.

\paragraph{Adversarial Attacks.}

Adversarial attack approaches are typically classified as either \textit{black-box} or \textit{white-box}, depending on whether the information about an attacked model is available. Black-box approaches~\cite{brendel2017decision, bhagoji2018practical, su2019one} may compensate a lack of information with an extensive computation. Since we consider high-resolution images and numerous clicks, the amount of computations required in a black-box adversarial attack is unfeasible. Accordingly, we are restricted with white-box approaches. 

The robustness of the conventional segmentation approaches was already explored~\cite{kamann2020benchmarking}; yet, as user inputs are not involved, the robustness could only be measured w.r.t image perturbations. A recent series of works~\cite{guan2023badsam, zhang2023attack, qiao2023robustness, wang2023empirical} measuring the robustness of SAM also focused on perturbing images rather than user prompts. In contrast, we fix input images and investigate the robustness w.r.t user inputs. We propose a fully differentiable white-box attack for generating adversarial user inputs, and formulate robustness metrics accordingly.

\section{TETRIS}

In this section, we briefly describe our self-collected dataset serving as a basis of the robustness benchmark. By creating TETRIS, we focused on usability, which implies proper licensing, no privacy violation (all depicted people gave an articulated consent), and avoiding other issues that may limit or forbid using a dataset. 

\subsection{Object Classes}

Object classes for TETRIS are chosen according to the task-specific requirements. Specifically, we select object classes seeming useful for image editing and labeling. Since we are unaware of any previous research dedicated to image editing scenarios, we cannot rely on a real-life object class distribution. So, we include classes present in PASCAL VOC 2012 \cite{pascal-voc-2012} and some other common classes from COCO \cite{coco}. Overall, we consider 9 metaclasses: \textit{transport, wild animal, object, domestic animal, food, architecture, plant, statue} are represented in TETRIS-\textsc{things}, while TETRIS-\textsc{people} contains only images of \textit{people}.

\begin{figure*}
    \centering
    \includegraphics[width=1.0\textwidth]{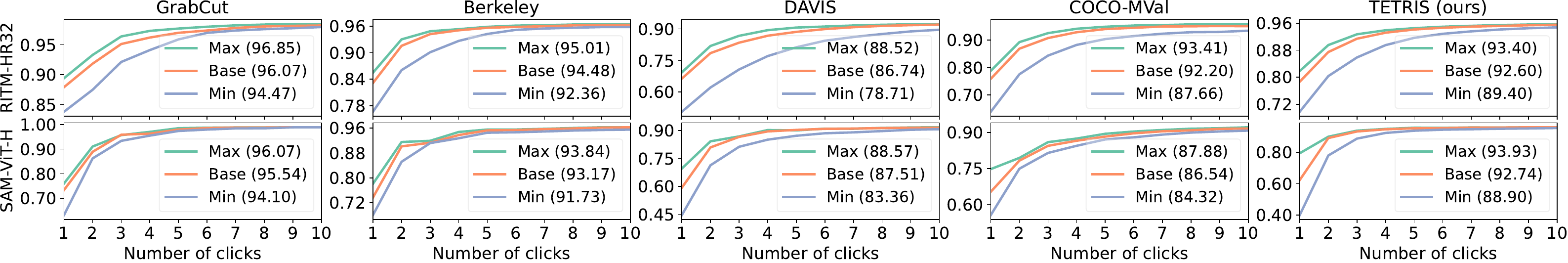}
    \caption{The minimizing, baseline, and maximizing trajectories of IoU for RITM and SAM models. Aggregated values (IoU-AuC) are given in brackets. 
    }
    \label{fig:min-base-max1}
\end{figure*}

\subsection{Image Acquisition and Annotation}

For TETRIS-\textsc{things}, we manually selected 1000 images from Unsplash\footnote{\url{https://unsplash.com/license}}. For TETRIS-\textsc{people}, we purchased 1000 photos directly from a crowdsourcing vendor. Age, gender, country, and race according to \cite{karkkainenfairface} were indicated by participants themselves. They also gave a mandatory consent to using their personal data and images; each user could donate from 1 to 5 photos. We restricted the photo resolution with at least of 2MP to ensure good quality of images.

The acquired images were segmented into polygonal regions using the CVAT \cite{cvat} labeling tool, and each such region was marked either as a foreground, a background, or an uncertain region. Next, we apply matting~\cite{park2022matteformer} to uncertain regions, so that they turned into either foreground or background. For images with several objects, we merged the corresponding alpha maps and thresholded them to obtain a binary segmentation mask. Finally, all masks were verified by expert annotators.

\begin{figure}[b]
    \centering
    \includegraphics[width=1.0\columnwidth]{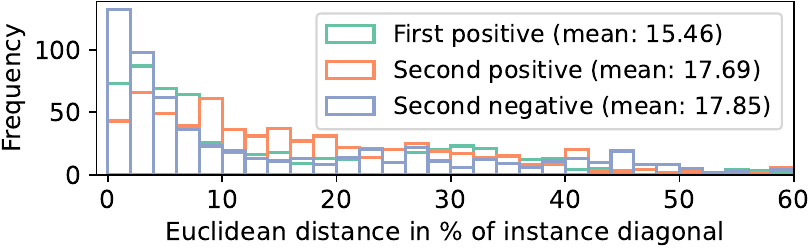}
    \caption{Distances between 1800 clicks made by real users and the ones generated using the baseline strategy.}
    \label{fig:histogram-of-diff}
\end{figure}

\section{Exploring Robustness Issues}

Below, we present the results of our real user study with two interaction rounds. Additionally, we analyze how the prediction quality varies in different possible click positions.

\subsection{Real User Study}

We selected five images per category from TETRIS-\textsc{things} for a real user study. We used a crowdsourcing web annotation platform, and asked hundreds of users to label images with a simple annotation tool.

\paragraph{First interaction round.}

Each performer was exposed with 1) a source image, and 2) the same image, overlapped with a predicted semi-transparent mask (or a ground-truth mask in the first interaction round). In the first interaction round, we asked the annotators to put a single click on the target object. In the first round, only positive clicks are allowed. The total of 600 users participated, resulting in 15 interactions per each of 40 images.

\paragraph{Second interaction round.} 

After completing the first round, the acquired user clicks were processed with RITM HRNet18 \cite{ritm}, and false positive and false negative per-pixel errors were calculated. For each type, we selected 40 samples with the largest error values. For clicks with the largest False Positive error, where the model predicted excessive masks, annotators were asked to make a negative click to exclude the redundant regions. Vice versa, for clicks with the largest False Negative error, users were requested to make the second positive click to cover missing areas. Other 1200 users were recruited in this round, providing 15 interactions per image.

\subsection{Exhaustive Search}

We also investigate the model quality in a full search over all integer input positions. Being arguably the simplest and the most intuitive way to measure the quality change w.r.t perturbed user inputs, this approach is resource-exhaustive: the number of forward passes grows linearly with both image height and width, making it impossible to evaluate on a full dataset in reasonable time.

We perform a brute-force on a pixel-wise grid and visualize the obtained results as heatmaps in Figure~\ref{fig:user-study-visual} (bottom rows). For each pixel, the color represents the IoU score obtained if clicking on this pixel; warmer hues mark higher IoU scores. Brute-force for a single $1024\times1024$ image takes over 4 hours using a single NVIDIA Tesla V100 to proceed, which encourages us to seek a faster approach for robustness evaluation, described below.

\begin{figure*}[ht]
    \centering
    \includegraphics[width=1.0\textwidth]{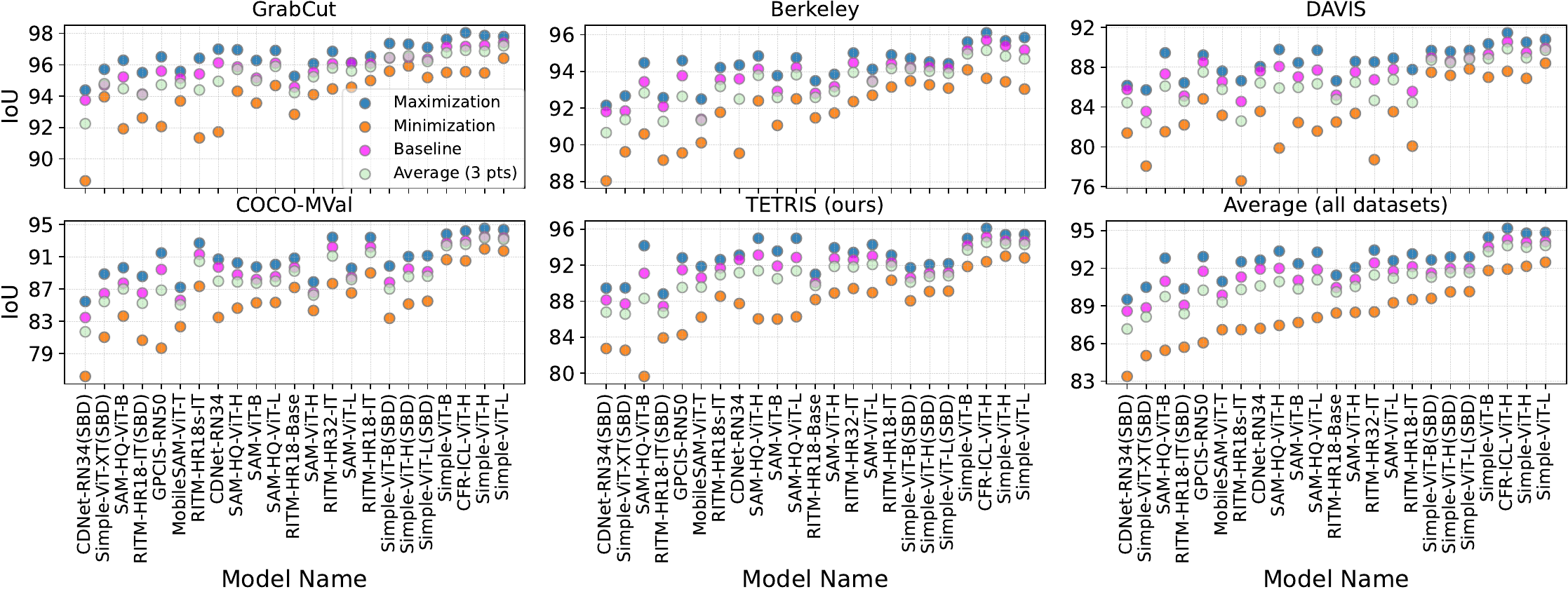}
    \caption{Visualization of IoU bounds of tested models. Columns are sorted by an average IoU-Min. As can be seen, there exists a dependency between the robustness (IoU-D, a delta between IoU-Max and IoU-Min) and the prediction quality.}
    \label{fig:min-base-max2}
\end{figure*}

\subsection{Analysis}

We observe that the click position obtained with the baseline clicking strategy is consistent with the real user click only in case of convex objects of simple shapes (e.g., a pizza). For more complex geometries, users tend to click in different areas of density, or salience objects' parts. Figure~\ref{fig:histogram-of-diff} shows distances between each user click and the click generated by the baseline strategy; all distances are normalized by an instance size (a diagonal length) for fair comparison between different objects. The error exceeds 15 percent on average, reaching a half of an instance diagonal size in some cases.

Nevertheless, user inputs are mostly gathered in a vicinity of the object's ``center'' (being understood subjectively based on a common sense rather than formal criteria), which might not actually coincide with the point being the furthest from the boundaries. The divergence of real and generated user clicks is especially tangible in case of long, thin objects, i.e., a snake. Besides, we notice that the quality of the tested model may variate within a large range depending on a click position. As one can see in Figure~\ref{fig:user-study-std}, the click position significantly affects the quality in more than on a half test samples. It also shows that users might easily unintentionally place clicks in such adversarial points, providing an unexpectedly low segmentation quality. 

\section{Proposed Evaluation Protocol}

Let us informally define a \textit{robustness} of an interactive segmentation model as its ability to output the same mask for any valid user input pointing to the same object. We consider only valid inputs: positive clicks must be placed within a yet unselected area of an object mask, while negative clicks should locate in some selected area outside the object mask.

\subsection{Adversarial Inputs}

To reduce the processing time from hours to seconds, compared to brute-force, we need to restrict the search space in a sensible way. We assume the center of the object to be a reasonable starting point, and then search for a local extreme in its vicinity.

\begin{figure}[b!]
    \centering
    \includegraphics[width=1.0\columnwidth]{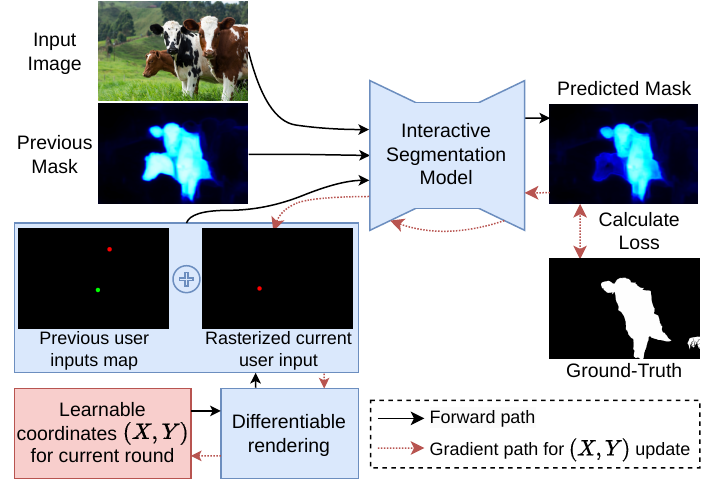}
    \caption{Overview of the proposed adversarial inputs generation. For models with raw $(X,Y)$ coordinates, such as \cite{kirillov2023segment}, we use a differentiable rendering step only for interaction location loss term.}
    \label{fig:pipeline}
\end{figure}

We select click positions with a white-box targeted attack. The overall scheme of our method is shown in Figure~\ref{fig:pipeline}. Using differentiable rendering \cite{xu2022live}, we encode clicks as maps with disks of a fixed radius marking click positions (since most models accept user inputs in this form). The radius is a hyperparameter, depending on the architecture of an attacked model. We calculate loss between the predicted and ground truth mask and run a gradient update to optimize click positions according to the chosen strategy. We use two strategies: one aims to minimize IoU, another targets at maximizing it. Surprisingly, even finding local extrema using the gradient descent method, a change of a click position has a great impact on the final quality (Figure~\ref{fig:min-base-max1}).

\begin{table*}[t]
\fontsize{9pt}{11pt}\selectfont
\tabcolsep=2pt
\begin{tabular}{|c|c|c|l|ccc|l|ccc|l|ccc|l|ccc|l|ccc|ll}
\cline{1-3} \cline{5-7} \cline{9-11} \cline{13-15} \cline{17-19} \cline{21-23}
\multirow{3}{*}{Method}      & \multirow{3}{*}{Model} & \multirow{3}{*}{Data}                                                    &                   & \multicolumn{3}{c|}{GrabCut}                    &  & \multicolumn{3}{c|}{Berkeley}                   &  & \multicolumn{3}{c|}{DAVIS}                      &  & \multicolumn{3}{c|}{COCO-MVal}                  &  & \multicolumn{3}{c|}{TETRIS (ours)}              &                      &                      \\ \cline{5-7} \cline{9-11} \cline{13-15} \cline{17-19} \cline{21-23}
                             &                        &                                                                          &                   & \multicolumn{3}{c|}{IoU (AuC@10)}               &  & \multicolumn{3}{c|}{IoU (AuC@10)}               &  & \multicolumn{3}{c|}{IoU (AuC@10)}               &  & \multicolumn{3}{c|}{IoU (AuC@10)}               &  & \multicolumn{3}{c|}{IoU (AuC@10)}               &                      &                      \\
                             &                        &                                                                          &                   & Min$\uparrow$            & Max$\uparrow$            & D$\downarrow$             &  & Min$\uparrow$            & Max$\uparrow$            & D$\downarrow$             &  & Min$\uparrow$            & Max$\uparrow$            & D$\downarrow$             &  & Min$\uparrow$            & Max$\uparrow$            & D$\downarrow$             &  & Min$\uparrow$            & Max$\uparrow$            & D$\downarrow$             &                      &                      \\ \cline{1-3} \cline{5-7} \cline{9-11} \cline{13-15} \cline{17-19} \cline{21-23}
MobileSAM                    & ViT-Tiny               & SA-1B                                                                    &                   & 93.69          & 95.58          & 1.89          &  & 90.11          & 92.49          & 2.38          &  & 83.16          & 87.60          & 4.44          &  & 82.32          & 87.20          & 4.88          &  & 86.22          & 91.85          & 5.64          &                      &                      \\ \cline{1-3} \cline{5-7} \cline{9-11} \cline{13-15} \cline{17-19} \cline{21-23}
\multirow{3}{*}{SAM}         & ViT-B                  & \multirow{3}{*}{SA-1B}                                                   &                   & 93.56          & 96.28          & 2.72          &  & 91.06          & 93.77          & 2.71          &  & 82.44          & 88.44          & 6.00          &  & 85.28          & 89.73          & 4.45          &  & 86.01          & 93.56          & 7.55          &                      &                      \\ \cline{2-2}
                             & ViT-L                  &                                                                          &                   & 94.57          & 96.13          & 1.57          &  & 92.70          & 94.12          & 1.41          &  & 83.53          & 88.90          & 5.37          &  & 86.51          & 89.57          & 3.06          &  & 88.94          & 94.28          & 5.33          &                      &                      \\ \cline{2-2}
                             & ViT-H                  &                                                                          &                   & 94.10          & 96.07          & 1.97          &  & 91.73          & 93.84          & 2.11          &  & 83.36          & 88.57          & 5.21          &  & 84.32          & 87.88          & 3.56          &  & 88.90          & 93.93          & 5.04          &                      &                      \\ \cline{1-3} \cline{5-7} \cline{9-11} \cline{13-15} \cline{17-19} \cline{21-23}
\multirow{3}{*}{SAM-HQ}      & ViT-B                  & \multirow{3}{*}{\begin{tabular}[c]{@{}c@{}}SA-1B\\ +44K\end{tabular}} &                   & 91.92          & 96.29          & 4.37          &  & 90.59          & 94.47          & 3.88          &  & 81.52          & 89.44          & 7.92          &  & 83.62          & 89.64          & 6.02          &  & 79.63          & 94.16          & 14.53         &                      &                      \\ \cline{2-2}
                             & ViT-L                  &                                                                          &                   & 94.68          & 96.91          & 2.23          &  & 92.51          & 94.74          & 2.24          &  & 81.57          & 89.69          & 8.12          &  & 85.32          & 90.04          & 4.72          &  & 86.26          & 94.98          & 8.72          &                      &                      \\ \cline{2-2}
                             & ViT-H                  &                                                                          &                   & 94.31          & 96.96          & 2.65          &  & 92.40          & 94.84          & 2.45          &  & 79.87          & 89.77          & 9.90          &  & 84.62          & 90.24          & 5.62          &  & 86.03          & 94.98          & 8.96          &                      &                      \\ \cline{1-3} \cline{5-7} \cline{9-11} \cline{13-15} \cline{17-19} \cline{21-23}
\multirow{2}{*}{CDNet}       & RN34                   & C+L                                                                      &                   & 91.72          & 96.99          & 5.27          &  & 89.54          & 94.35          & 4.81          &  & 83.56          & 88.06          & 4.50          &  & 83.46          & 90.70          & 7.24          &  & 87.72          & 93.12          & 5.39          &                      &                      \\ \cline{2-3}
                             & RN34                   & SBD                                                                      &                   & 88.60          & 94.39          & 5.79          &  & 88.04          & 92.15          & 4.11          &  & 81.39          & 86.15          & 4.77          &  & 76.14          & 85.44          & 9.30          &  & 82.73          & 89.44          & 6.71          & \multicolumn{1}{c}{} &                      \\ \cline{1-3} \cline{5-7} \cline{9-11} \cline{13-15} \cline{17-19} \cline{21-23}
GPCIS                        & RN50                   & C+L                                                                      &                   & 92.06          & 96.51          & 4.45          &  & 89.56          & 94.59          & 5.03          &  & 84.81          & 89.21          & 4.41          &  & 79.65          & 91.47          & 11.83         &  & 84.26          & 92.81          & 8.56          &                      &                      \\ \cline{1-3} \cline{5-7} \cline{9-11} \cline{13-15} \cline{17-19} \cline{21-23}
\multirow{5}{*}{RITM}        & HR18s-IT               & \multirow{4}{*}{C+L}                                                     &                   & 91.34          & 96.42          & 5.08          &  & 91.77          & 94.21          & 2.44          &  & 76.58          & 86.63          & 10.06         &  & 87.32          & 92.69          & 5.37          &  & 88.53          & 92.60          & 4.08          &                      &                      \\ \cline{2-2}
                             & HR18                   &                                                                          &                   & 92.84          & 95.27          & 2.43          &  & 91.47          & 93.49          & 2.01          &  & 82.48          & 86.62          & 4.14          &  & 87.18          & 90.82          & 3.64          &  & 88.17          & 90.99          & 2.82          & \multicolumn{1}{c}{} & \multicolumn{1}{c}{} \\ \cline{2-2}
                             & HR18-IT                &                                                                          &                   & 95.00          & 96.54          & 1.54          &  & 93.15          & 94.90          & 1.75          &  & 80.06          & 87.76          & 7.69          &  & 88.98          & 93.39          & 4.40          &  & 90.32          & 93.11          & 2.79          & \multicolumn{1}{c}{} & \multicolumn{1}{c}{} \\ \cline{2-2}
                             & HR32-IT                &                                                                          &                   & 94.47          & 96.85          & 2.39          &  & 92.36          & 95.01          & 2.65          &  & 78.71          & 88.52          & 9.81          &  & 87.66          & 93.41          & 5.74          &  & 89.40          & 93.40          & 4.00          & \multicolumn{1}{c}{} & \multicolumn{1}{c}{} \\ \cline{2-3} \cline{5-7} \cline{9-11} \cline{13-15} \cline{17-19} \cline{21-23}
                             & HR18-IT                & SBD                                                                      &                   & 92.62          & 95.50          & 2.88          &  & 89.17          & 92.57          & 3.40          &  & 82.20          & 86.43          & 4.22          &  & 80.62          & 88.56          & 7.93          &  & 83.91          & 88.80          & 4.89          & \multicolumn{1}{c}{} &                      \\ \cline{1-3} \cline{5-7} \cline{9-11} \cline{13-15} \cline{17-19} \cline{21-23}
\multirow{7}{*}{SimpleClick} & ViT-B                  & \multirow{3}{*}{C+L}                                                     &                   & 95.51          & 97.63          & 2.12          &  & \textbf{94.09} & 95.61          & 1.51          &  & 86.98          & 90.34          & 3.36          &  & 90.64          & 93.80          & 3.17          &  & 91.85          & 94.97          & 3.12          &                      &                      \\ \cline{2-2}
                             & ViT-L                  &                                                                          &                   & \textbf{96.41} & 97.80          & \textbf{1.39} &  & 93.04          & \underline{95.85}    & 2.81          &  & \textbf{88.40} & \underline{90.79}    & 2.39    &  & \underline{91.73}    & \underline{94.37}    & \underline{2.64}    &  & \underline{92.81}    &\underline{95.42}    & \underline{2.60}    & \multicolumn{1}{c}{} & \multicolumn{1}{c}{} \\ \cline{2-2}
                             & ViT-H                  &                                                                          &                   & 95.48          & \underline{97.87}    & 2.39          &  & 93.44          & 95.66          & 2.23          &  & 86.88          & 90.50          & 3.62          &  & \textbf{91.97} & \textbf{94.53} & \textbf{2.56} &  & \textbf{92.99} & 95.38          & \textbf{2.39} & \multicolumn{1}{c}{} & \multicolumn{1}{c}{} \\ \cline{2-3} \cline{5-7} \cline{9-11} \cline{13-15} \cline{17-19} \cline{21-23}
                             & ViT-XT                 & \multirow{4}{*}{SBD}                                                     & \multirow{4}{*}{} & 93.96          & 95.72          & 1.76          &  & 89.62          & 92.66          & 3.05          &  & 78.06          & 85.71          & 7.65          &  & 81.01          & 88.86          & 7.85          &  & 82.53          & 89.47          & 6.94          &                      &                      \\ \cline{2-2}
                             & ViT-B                  &                                                                          &                   & 95.59          & 97.36          & 1.77          &  & 93.49          & 94.71          & \textbf{1.22} &  & 87.49          & 89.67          & \underline{2.19}          &  & 83.36          & 89.84          & 6.48          &  & 88.04          & 91.72          & 3.68          & \multicolumn{1}{c}{} & \multicolumn{1}{c}{} \\ \cline{2-2}
                             & ViT-L                  &                                                                          &                   & 95.19          & 97.10          & 1.91          &  & 93.09          & 94.42          & 1.33          &  & \underline{87.82}    & 89.68          & \textbf{1.86} &  & 85.47          & 91.12          & 5.65          &  & 89.10          & 92.16          & 3.06          & \multicolumn{1}{c}{} & \multicolumn{1}{c}{} \\ \cline{2-2}
                             & ViT-H                  &                                                                          &                   & \underline{95.92}    & 97.32          & \underline{1.39}    &  & 93.27          & 94.52          & \underline{1.25}          &  & 87.17          & 89.57          & 2.40          &  & 85.13          & 91.02          & 5.89          &  & 89.07          & 92.06          & 2.99          & \multicolumn{1}{c}{} & \multicolumn{1}{c}{} \\ \cline{1-3} \cline{5-7} \cline{9-11} \cline{13-15} \cline{17-19} \cline{21-23}
CFR-ICL                      & ViT-H                  & C+L                                                                      &                   & 95.56          & \textbf{98.04} & 2.48          &  & \underline{93.63}    & \textbf{96.11} & 2.48          &  & 87.58          & \textbf{91.45} & 3.87          &  & 90.49          & 94.19          & 3.71          &  & 92.38          & \textbf{96.09} & 3.72          &                      &                      \\ \cline{1-3} \cline{5-7} \cline{9-11} \cline{13-15} \cline{17-19} \cline{21-23}
\end{tabular}
\caption{The quality and robustness scores of different models, measured on the standard datasets and our novel TETRIS dataset. The best results are bold, the second best are underlined. SimpleClick and CFR-ICL are more robust than other tested approaches. Still, even state-of-the-art models are extremely sensitive to the positions of user clicks, which may cause an unstable performance in a real-world scenario.}
\label{tbl:base-d-og}
\end{table*}

\subsection{Proposed Metrics}

\begin{figure}[b!]
    \centering
    \includegraphics[width=1.0\columnwidth]{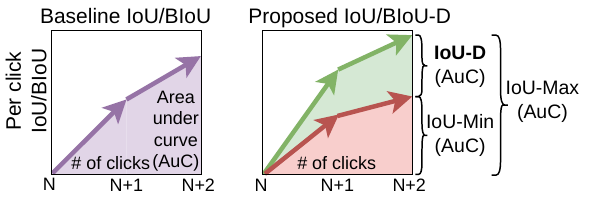}
    \caption{Left: the standard IoU/BIoU-AuC score. Right: the proposed IoU/BIoU-D robustness score. We consider 10 clicks and normalize the area to be within $[0, 1]$.}
        \label{fig:second_metric}
\end{figure}

Sequentially optimizing each interaction, we obtain two click trajectories, referred to as the \textit{minimizing trajectory} and \textit{maximizing trajectory}. Similarly, we address the click trajectory obtained via the baseline clicking strategy, as the \textit{baseline trajectory}. Based on the obtained trajectories, we propose a robustness metric specifically for the task of interactive segmentation (an intuitive explanation is given in Figure~\ref{fig:second_metric}).

\noindent\textbf{IoU/BIoU-Min/Max} — the area under the minimizing/maximizing trajectory curve, the quality metric on a generated trajectory of the worst/best adversarial clicks.

\noindent\textbf{IoU/BIoU-D} — the difference between the area under curves of maximizing and minimizing trajectories. As depicted in Figure~\ref{fig:min-base-max1}, maximizing, minimizing, and baseline trajectories converge with an increasing number of clicks, and the accuracy gap decreases accordingly. Therefore, the difference between trajectories is the most divisible and hence informative for a few clicks; accordingly, we consider only 10 clicks in all our experiments. 

\subsection{Evaluation Setup}

\begin{figure*}[ht]
    \centering
    \includegraphics[width=1.0\textwidth]{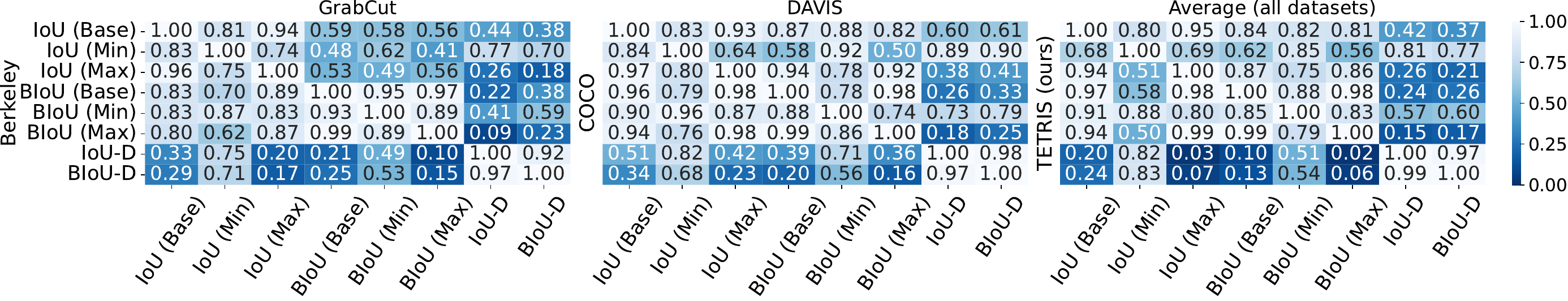}
    \caption{Cross-metric Spearman's rank correlations measured for different datasets.}
    \label{fig:cross-metric}
\end{figure*}

\begin{figure*}[ht]
    \centering
    \includegraphics[width=1.0\textwidth]{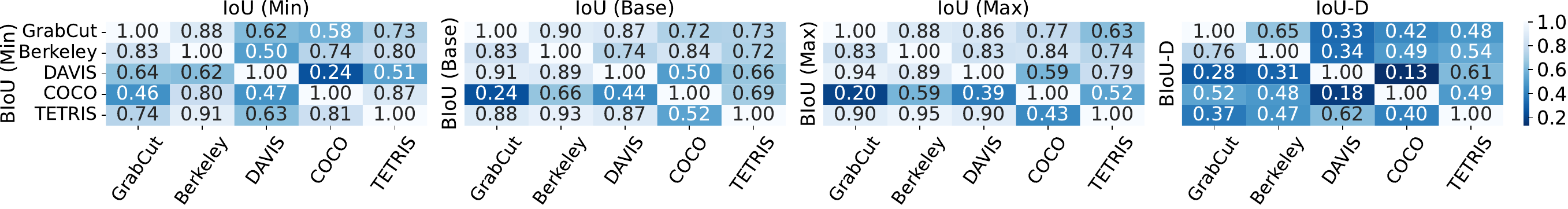}
    \caption{Cross-dataset Spearman's rank correlations of IoU and BIoU scores.}
    \label{fig:cross-dataset}
\end{figure*}

We narrow down our evaluation with recent methods having an open-source codebase: RITM \cite{ritm}, CDNet \cite{chen2021conditional}, SimpleClick \cite{liu2022simpleclick}, CFR-ICL \cite{sun2023cfricl}, and GPCIS \cite{zhou2023interactivegpcis}. Besides, we experiment with promptable interactive segmentation methods from the SAM family: the original SAM \cite{kirillov2023segment}, SAM-HQ \cite{ke2023segment}, MobileSAM \cite{zhang2023faster}. Overall, we validate 23 checkpoints on the 5 interactive segmentation datasets: GrabCut, Berkeley, DAVIS, and COCO-MVal, as well as on our novel TETRIS dataset. The obtained quality scores are listed in Table~\ref{tbl:base-d-og} and presented in a visual form with the baseline strategy in Figure~\ref{fig:min-base-max2}. 

We run no more than 10 optimization iterations to restrict the number of calculations. 
The gradient updates are calculated with an Adam optimizer~\cite{kingma2014adam}. To compare models with different input resolution fairly, we linearly scale the learning rate by an input size factor: $LR=\frac{5\sqrt{H^{2} + W^{2}}}{400\sqrt{2}}$, where $H, W$ denote image height and width in pixels, respectively. For minimizing and maximizing trajectories, the first iteration is selected with the baseline strategy and the consecutive clicks are placed greedily one by one.

Without any constraints, the maximization strategy yields points in between the object parts but outside the object mask. While providing the best quality, such click positions are unlikely to be made by a real user. Also, the minimizing optimization can easily converge outside an object of interest. Thus, to generate valid clicks in an adversarial optimization, we impose an additional constraint. We calculate a distance transform map for false positive and false negative areas. For a positive click, we sum up the distances in the false negative positions covered by a circle representing this click, for a negative click -- in the false positive positions, respectively. This gives an \textit{interaction location loss}. The total loss is a weighted sum of a Dice~\cite{dice1945measures} loss and an interaction location loss, scaled by 1000. During the optimization, we use the following update scheme:

\begin{enumerate}
    \item Initialize an optimizable click position according to the baseline strategy;
    \item Start the optimization by minimizing / maximizing a loss function;
    \item Accept the new generated click, if IoU decreases / increases and the \textit{interaction location loss} does not increase by more than 5\%: (for objects thinner than the click radius, the loss inevitably gets worse even with a precise click, so we allow a small margin);
    \item Save the predicted mask and click location from the best iteration, and use them in the next interaction round.
\end{enumerate}

For models that accept raw click coordinates, we directly optimize the coordinates and use the differentiable rendering step only to compute \textit{interaction location loss}. We follow the same evaluation procedure as is used in the original methods, applying ZoomIn~\cite{sofiiuk2020f}, Cascade-Refinement~\cite{sun2023cfricl}, test-time augmentation flips~\cite{ritm}, selecting a mask by a predicted score~\cite{kirillov2023segment}, etc.

\subsection{Discussion}

Based on the results of our user study (Figures~\ref{fig:user-study-visual}, and~\ref{fig:user-study-std}) and robustness evaluation (Table~\ref{tbl:base-d-og}; Figures~\ref{fig:min-base-max1}, and~\ref{fig:min-base-max2}), we can conclude that \textbf{state-of-the-art interactive segmentation models are extremely sensitive to the positions of user clicks}. 

An exhaustive search on a pixel grid reveals that clicking on some coordinates within an object may unexpectedly cause a significant accuracy drop. We further show that for few points obtained through an adversarial attack, the quality may fluctuate significantly even within a small homogeneous area. We attribute such undesired behavior to the model selecting a part of an object (like a single slice of pepperoni) rather than the entire object (like a pepperoni pizza). Since the formulation of the interactive segmentation task is naturally fuzzy, such an ambiguity occurring is inevitable to a certain extent. Besides, the complexity of an object's shape affects the model's performance greatly. However, when developing an interactive segmentation model, one should aim to minimize those effects. 

The minimizing, maximizing, and baseline trajectories do converge with an increasing number of clicks. The difference in quality is the largest during the first few interactions. This actually means that \textbf{clicking in any sensible way (i.e. committing only \textit{valid clicks}) will provide an acceptable result -- but, possibly, after many interactions}.

Furthermore, we explore pairwise correlations between quality and robustness metrics (Figure~\ref{fig:cross-metric}) and compare model rankings on different datasets (Figure~\ref{fig:cross-dataset}). It can be seen that:

\begin{itemize}
    \item IoU/BIoU-Base strongly correlates with IoU/BIoU-Max; therefore, \textbf{the baseline evaluation protocol implicitly ranks models by the best possible quality, but does not reflect their performance in the worst case} (Figure~\ref{fig:cross-metric}); 
    \item IoU/BIoU-D strongly correlates with IoU/BIoU-Min, while the correlations with IoU/BIoU-Max is weaker (Figure~\ref{fig:cross-metric}). We attribute this to the fact that most of the quality spread is associated with a performance drop of the minimizing trajectory. According to Figures~\ref{fig:min-base-max1}, and~\ref{fig:min-base-max2}, optimizing adversarial inputs for the maximizing trajectory is much more difficult than searching for the worst clicks;
    \item The model ranking turns out to be dataset-specific. The ranking on TETRIS differs from the ranking on low-resolution datasets (Figure~\ref{fig:cross-dataset}).
\end{itemize}

\section{Conclusion}

In this study, we showed that the prediction quality of click-based interactive segmentation models depends heavily on the click location. To this end, we conducted a real user study and analyzed 1800 participant responses. Guided by this empirical evidence, we proposed the adversarial input generation strategy, and formulated the robustness score, which is estimated based on multiple generated trajectories. We evaluated the robustness of dozens of open-sourced models on the well-known datasets; and also on the proposed TETRIS benchmark with 2000 high-resolution images manually labeled with fine segmentation masks. 


\bibliography{main}

\end{document}


\twocolumn[
\smaketitle
\vspace*{-2.55cm}
]


\section{Adversarial inputs}
\subsection{Selection number of iterations}

\begin{figure}[h]
    \centering
    \includegraphics[width=1.0\columnwidth]{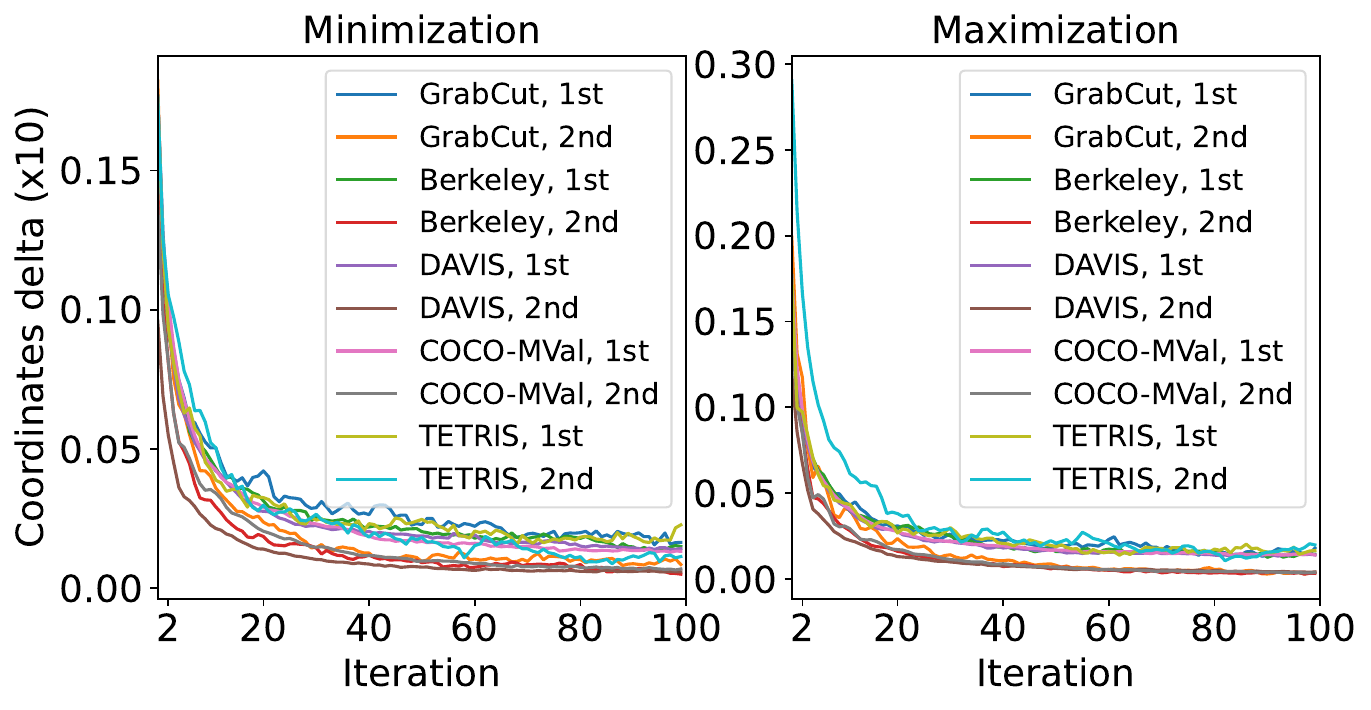}
    \caption{Averaged (for all samples in dataset) Euclidean distance of coordinates in neighboring optimization iterations for 1st and 2nd clicks. Linear coordinates are normalized to the maximum dimension of the image. As can be seen, the main trend for both clicks and for the optimization mode behaves similarly.}
        \label{fig:deltas_per_step}
\end{figure}

To select the number of optimization steps, we run 101 optimization steps of RITM-HRNet18-IT~\cite{ritm} and explore the pairwise coordinates deltas (Figure~\ref{fig:deltas_per_step}) of neighboring optimization steps click coordinates. The first few iterations are the most impactful, while the effect of each consequent step gradually deteriorates. Thereby, we limit the number of iterations with 10 to significantly reduce the number of calculations. Thus, single input optimization requires about 10s on NVIDIA Tesla V100 with $400\times400$ input size. Time may vary depending on the model and input size.

\subsection{Evaluation}

We narrow down our evaluation with recent methods having an open-source codebase, namely, RITM \cite{ritm}, CDNet \cite{chen2021conditional}, SimpleClick \cite{liu2022simpleclick}, CFR-ICL \cite{sun2023cfricl}, and GPCIS \cite{zhou2023interactivegpcis} as well as promptable interactive segmentation methods from the SAM family: the original SAM \cite{kirillov2023segment}, SAM-HQ \cite{ke2023segment}, MobileSAM \cite{zhang2023faster}. We carefully implemented interactive segmentation evaluation pipeline, since the original code presented only by a demo. For these models, rasterization of user inputs occurs only in the regularizing term (i.e. \textit{interaction location loss}), however the raw coordinates are fed into the model.

Evaluation results for BIoU metric presented in Table~\ref{tbl:base-d-og} and in graphical form in Figure~\ref{fig:biou_bounds}. Visualization of \textit{minimization, baseline, maximization} trajectories provided in Figure~\ref{fig:plot1}. Only the largest models of each type presented for the best view.

Besides, we provide visualizations of the adversarial clicks obtained using our optimization method for RITM~\cite{ritm} (in Figure~\ref{fig:traj1}) and Segment Anything~\cite{kirillov2023segment} (in Figure~\ref{fig:traj2}).

\section{Exploring Robustness Issues}

\subsection{First interaction round}

We provide visualizations of all real-user clicks as well as full brute-force on an integer coordinate grid and visualized the obtained results as heatmaps in Figure~\ref{fig:supplem_user_study1}, Figure~\ref{fig:supplem_user_study2}, Figure~\ref{fig:supplem_user_study3}, Figure~\ref{fig:supplem_user_study4} for RITM~\cite{ritm}, SimpleCick~\cite{liu2022simpleclick} and Segment Anything~\cite{kirillov2023segment} models. For each pixel, the color represents the IoU/BIoU score obtained from clicking on this pixel; warmer hues mark higher IoU/BIoU scores. In Figure~\ref{fig:user-study-std11}, Figure~\ref{fig:user-study-std12} we provide numerical bounds for IoU and BIoU metric values.

\subsection{Second interaction round}

A complete brute-force of even two sequential clicks is computationally impossible in reasonable time and requires the $(H \times W) ^ 2$ forward passes for image with $H \times W$ shape, so we do not provide all two-clicks combination results. However, we provide visualizations of all real-user second positive clicks in Figure~\ref{fig:second_positive}, and second negative in Figure~\ref{fig:second_negative}. In Figure~\ref{fig:user-study-std21}, Figure~\ref{fig:user-study-std22}, Figure~\ref{fig:user-study-std31}, Figure~\ref{fig:user-study-std32} provided numerical bounds for IoU and BIoU metric values on second positive and second negative clicks respectively.

This study demonstrates that the spread of quality after a second negative click is slightly less than after a second positive one, even the distance from the \textit{baseline} click is greater.

\begin{figure*}[ht]
    \begin{subfigure}[t]{0.245\textwidth}
        \includegraphics[width=\textwidth]{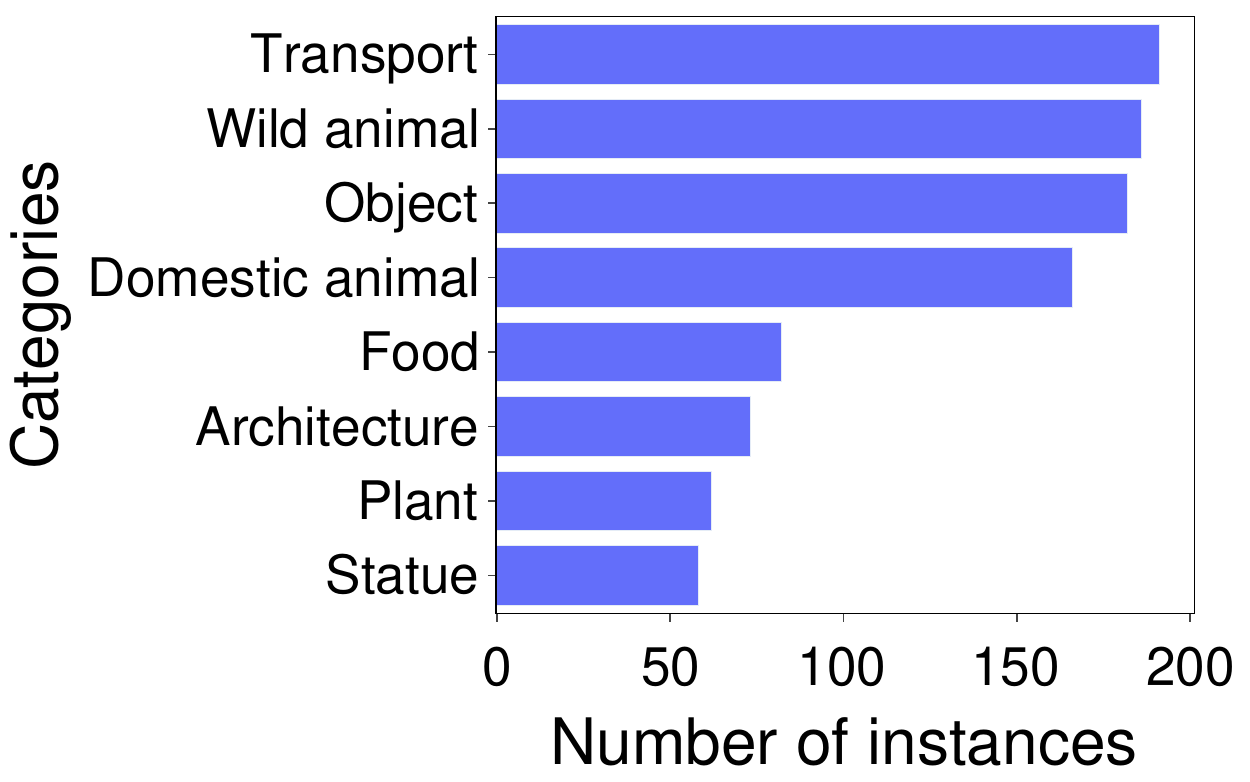}
        \caption{Number of images per category (except ``people'' category).}\label{fig:num-instances}
    \end{subfigure}
    \hfill
    \begin{subfigure}[t]{0.245\textwidth}
        \includegraphics[width=\linewidth]{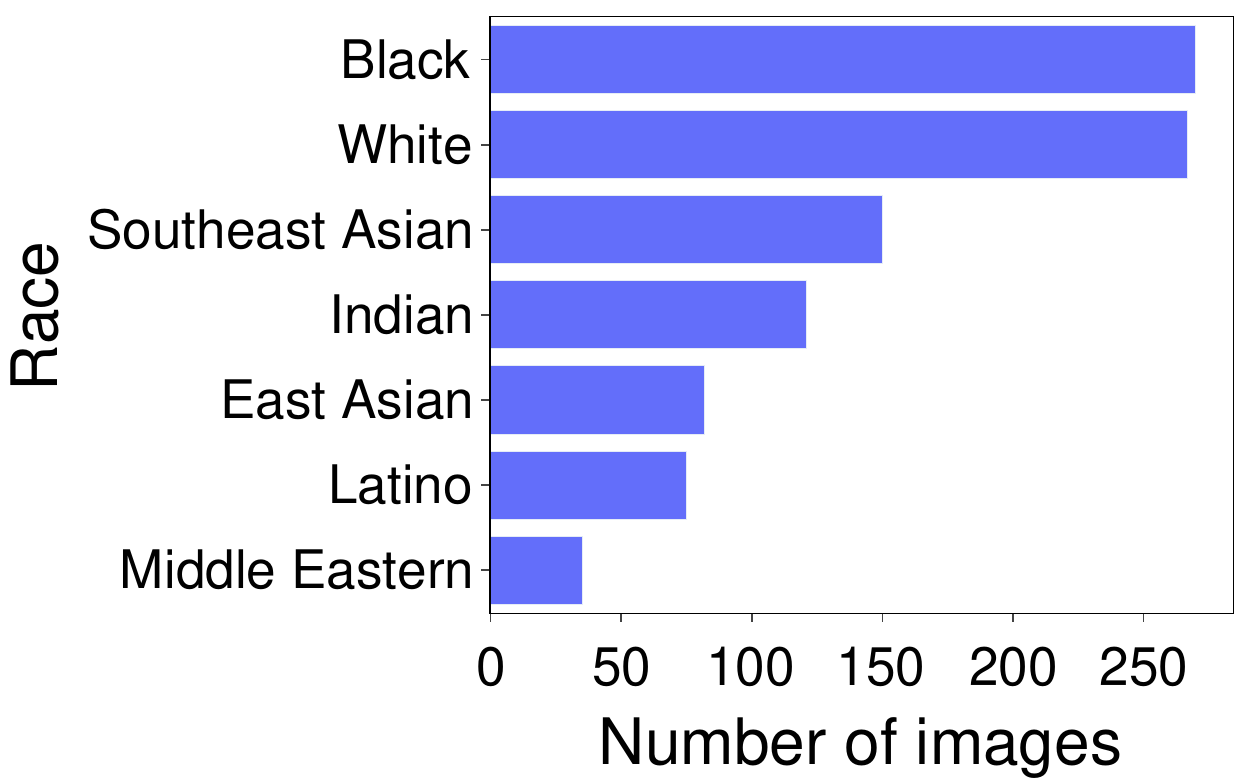}
        \caption{Distribution of races.}\label{fig:races}
    \end{subfigure}
    \hfill
    \begin{subfigure}[t]{0.245\textwidth}
        \includegraphics[width=\linewidth]{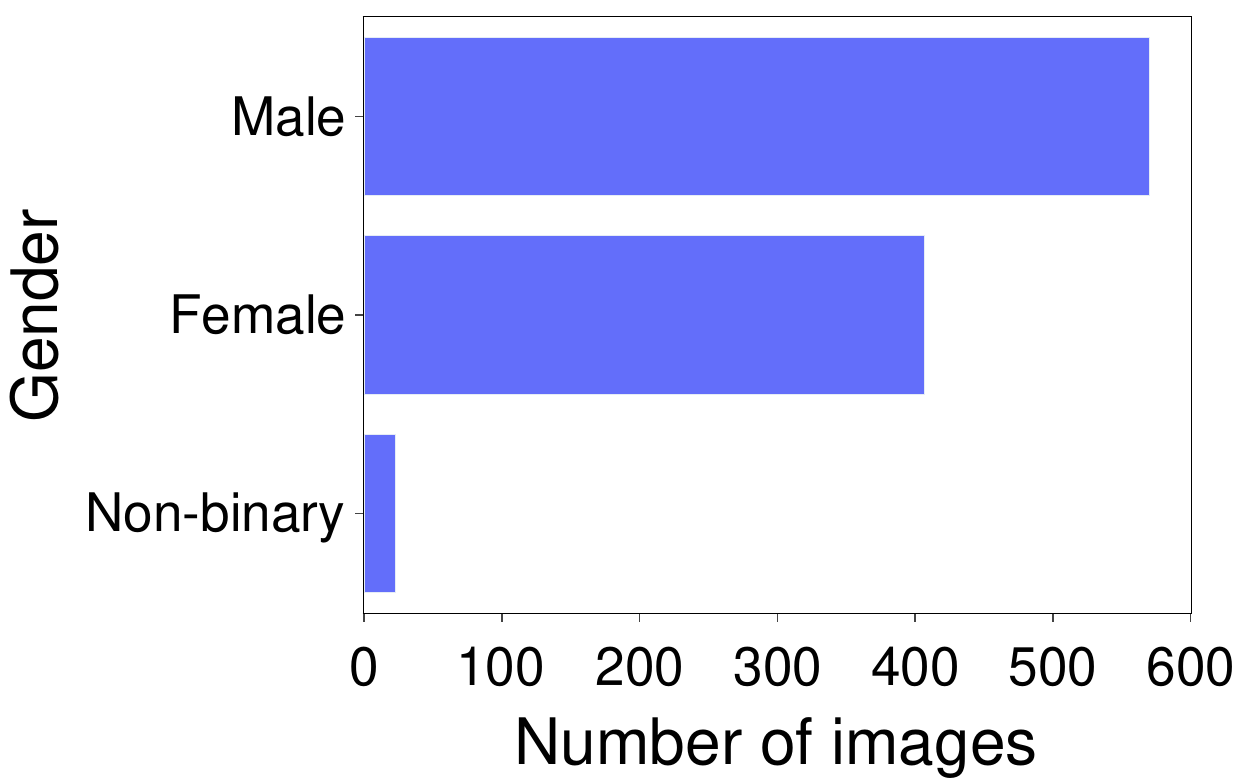}
        \caption{Gender distribution.}\label{fig:gender}
    \end{subfigure}
    \hfill
    \begin{subfigure}[t]{0.245\textwidth}
        \includegraphics[width=\linewidth]{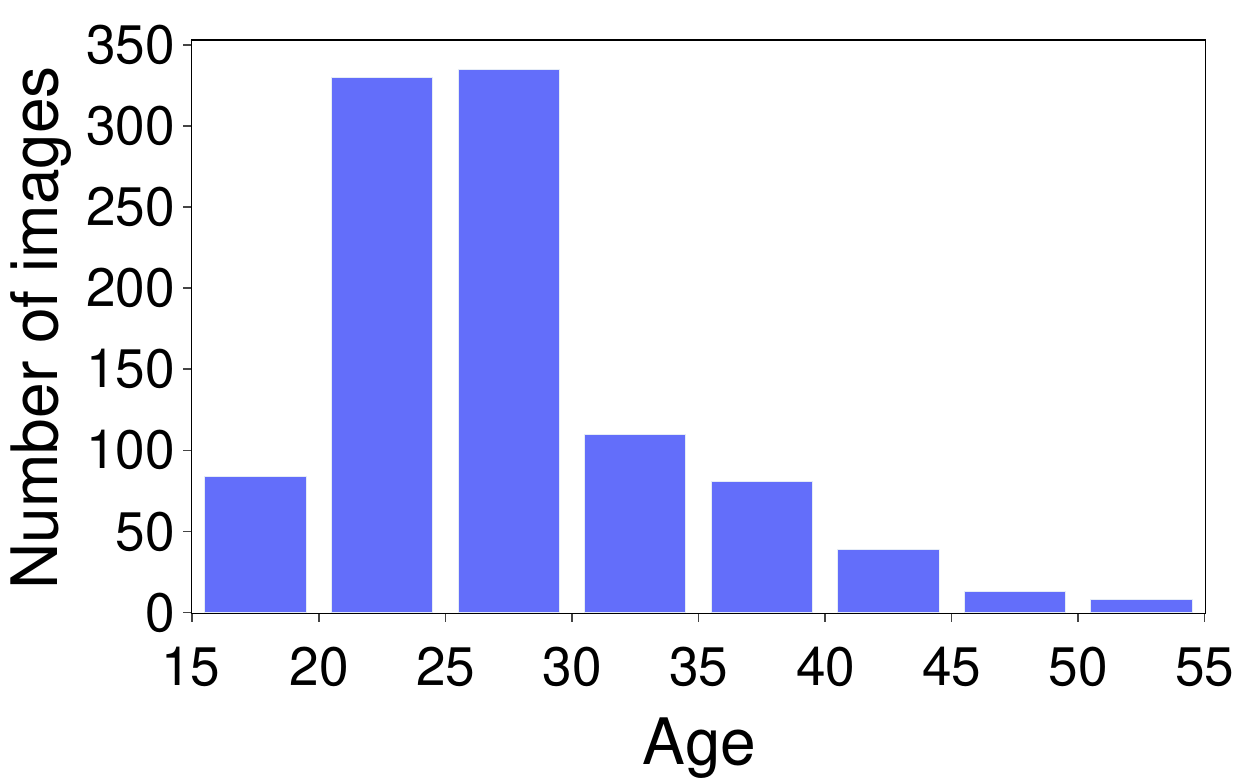}
        \caption{Distribution of age groups.}\label{fig:age}
    \end{subfigure}
    \hfill
    \begin{subfigure}[t]{0.245\textwidth}
        \includegraphics[width=\linewidth]{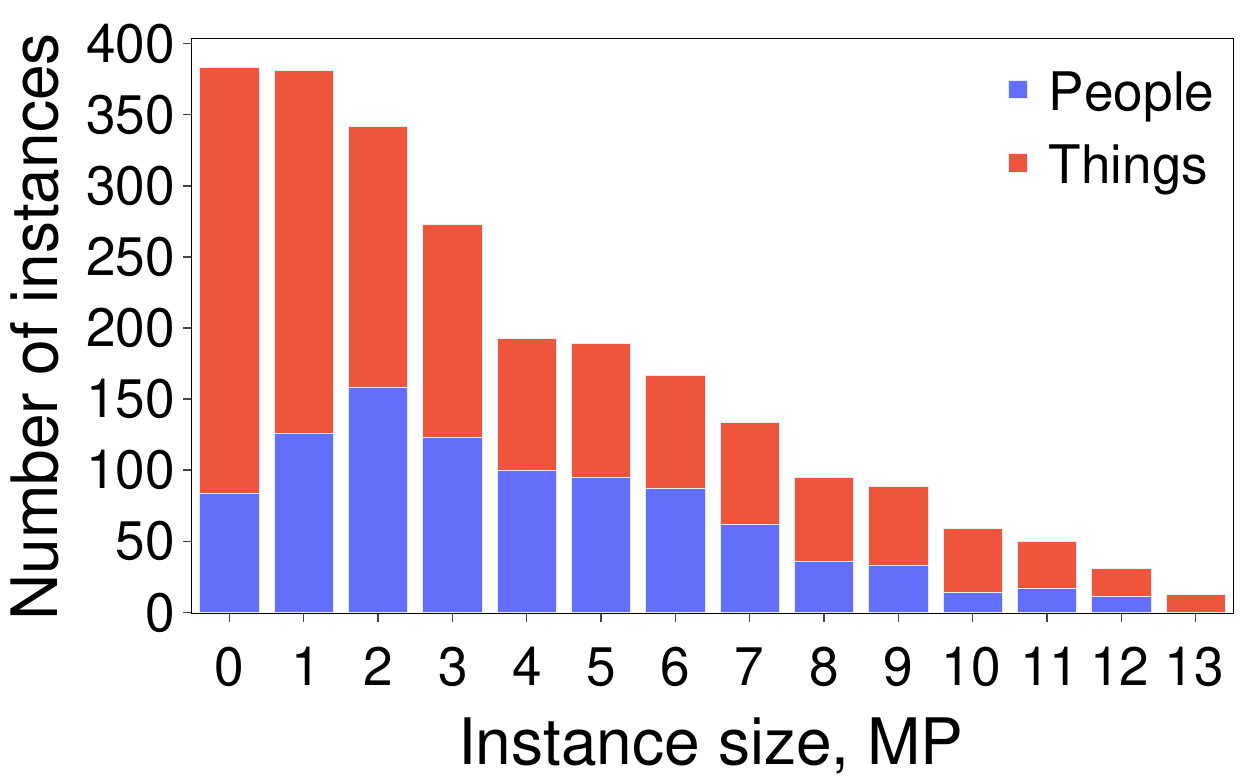}
            \caption{Instance sizes distribution.}
            \label{fig:instance-size}
    \end{subfigure}
    \hfill
    \begin{subfigure}[t]{0.245\textwidth}
        \includegraphics[width=\linewidth]{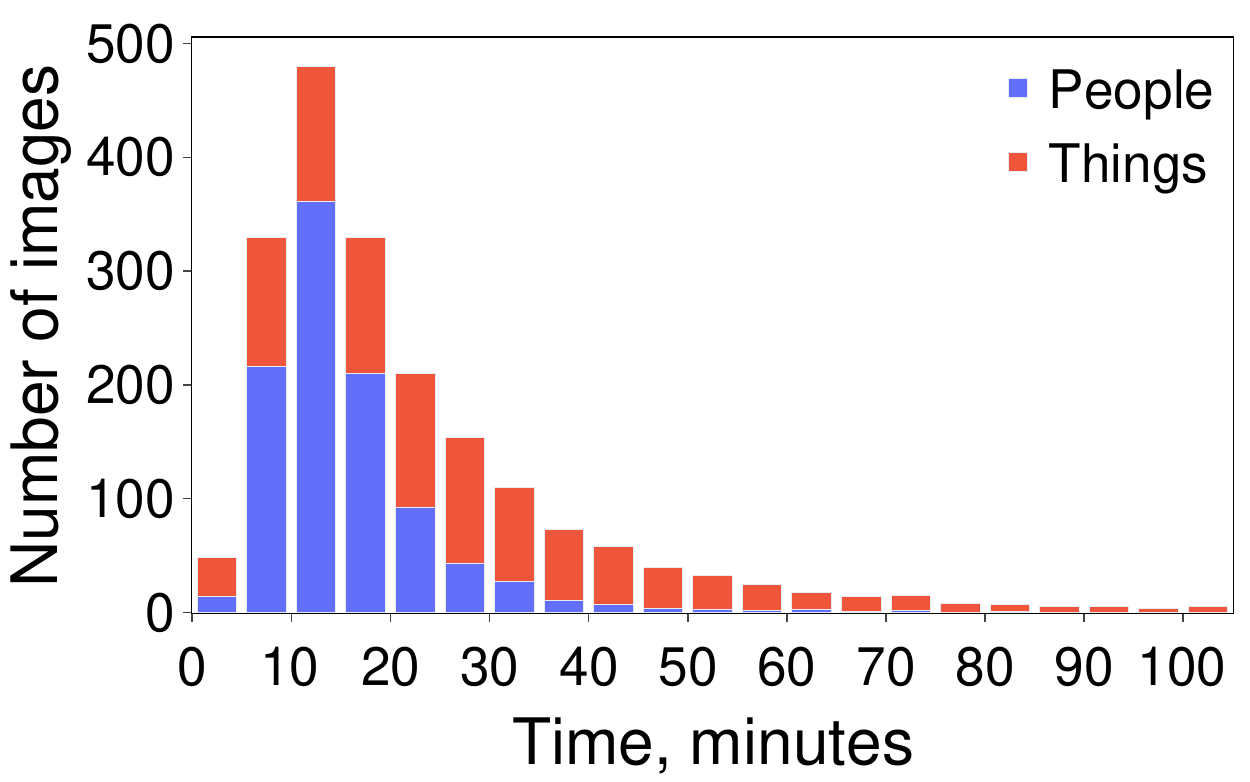}
            \caption{Annotation time per image. Excluded the most difficult image with 5+ hours to complete.}
            \label{fig:annotation-time}
    \end{subfigure}
    \hfill
    \begin{subfigure}[t]{0.305\textwidth}
        \includegraphics[width=\linewidth]{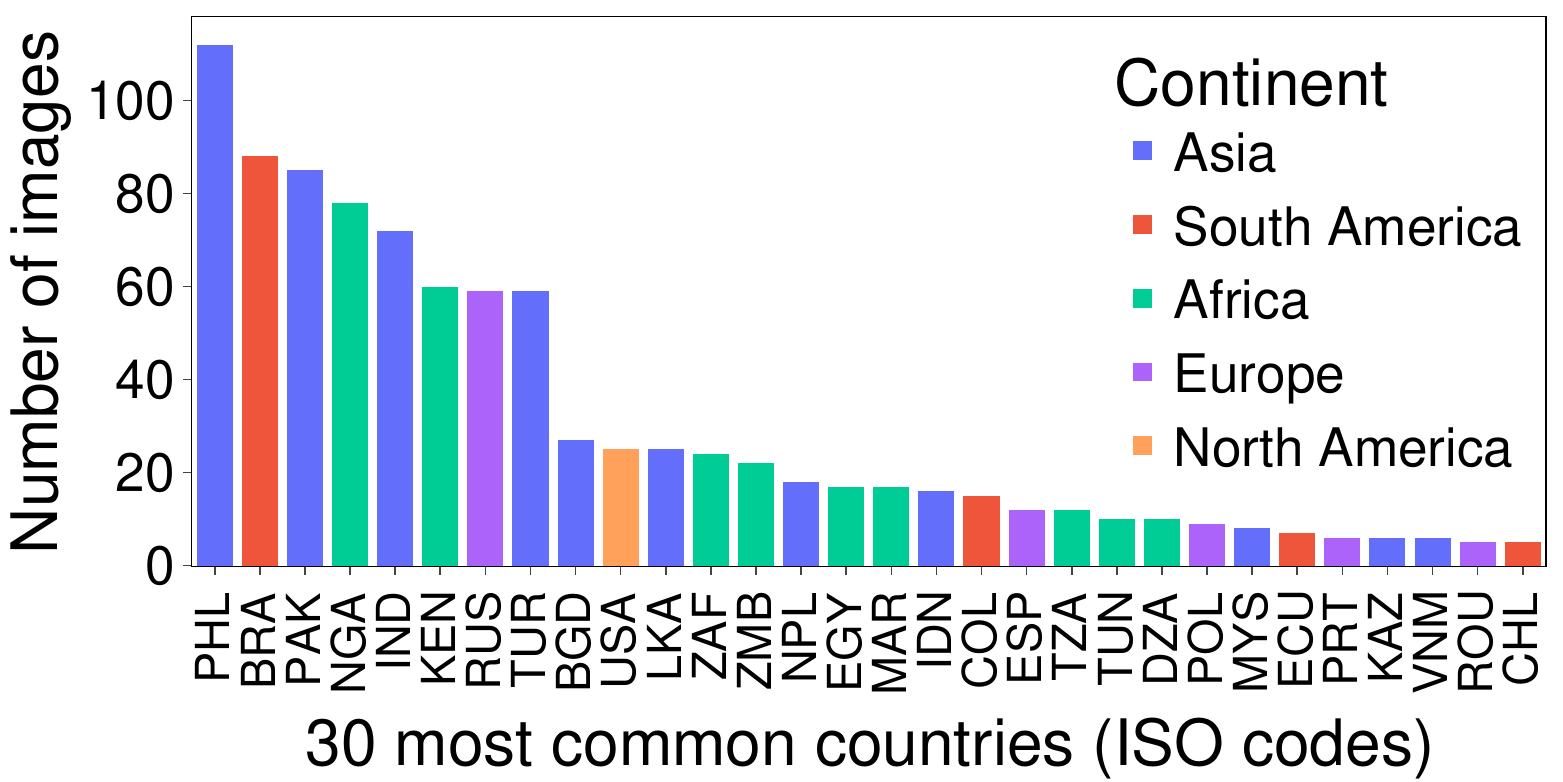}
        \caption{Geographical distribution.}\label{fig:geo}
    \end{subfigure}
    \hfill
    \begin{subfigure}[t]{0.185\textwidth}
        \includegraphics[width=\linewidth]{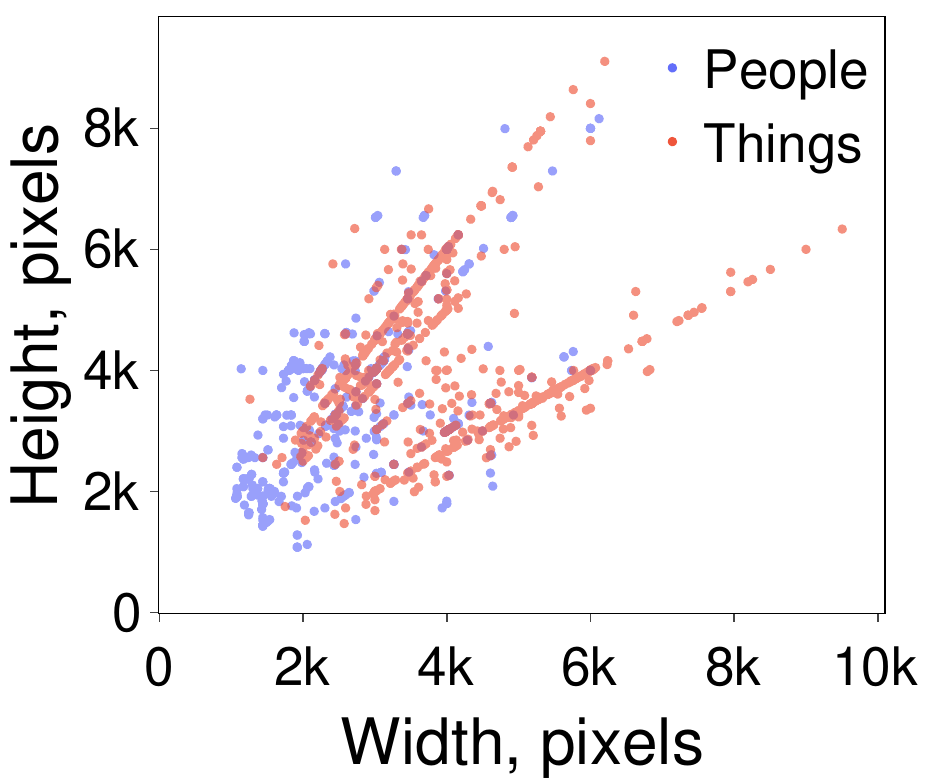}
        \caption{Image heights and widths.}
        \label{fig:resolution-plots}
    \end{subfigure}

    \caption{Statistics of TETRIS dataset.}\label{fig:stats}
    
\end{figure*}

\begin{table}[t!]
    \centering
    \begin{tabular}{|l|c|c|c|}
    \hline
    Dataset       & \begin{tabular}[c]{@{}c@{}}Images\\ (count)\end{tabular} & \begin{tabular}[c]{@{}c@{}}Instances\\ (count)\end{tabular} & \begin{tabular}[c]{@{}c@{}}Median \\ resolution\\ (MP)\end{tabular} \\ \hline
    GrabCut       & 50                                                       & 50                                                          & 0.25                                                                \\ \hline
    Berkeley      & 96                                                       & 100                                                         & 0.15                                                                \\ \hline
    SBD           & \textbf{2857}                                                     & \textbf{6671}                                                        & 0.2                                                                 \\ \hline
    PASCAL VOC    & 1449                                                     & 3417                                                        & 0.2                                                                 \\ \hline
    DAVIS         & 345                                                      & 345                                                         & 0.4                                                                 \\ \hline
    Composition1K & 1000                                                     & 1000                                                        & 2                                                                   \\ \hline
    BIG           & 150                                                      & 150                                                         & 9                                                                   \\ \hline
    AIM500        & 500                                                      & 500                                                         & 2                                                                   \\ \hline
    AM-2k         & 200                                                     & 200                                                        & 2                                                                   \\ \hline
    PM-10k        & 1000                                                    & 1000                                                       & 2                                                                   \\ \hline
    TETRIS (ours) & 2000                                                     & 2531                                                        & \textbf{12}                                                                  \\ \hline
    \end{tabular}
    \caption{Comparison of TETRIS benchmark with the existing ones (or validation subsets for datasets without test split). Our benchmark contains images and high-quality masks at notably larger resolution than existing datasets.}
    \label{tbl:is-datasets}
\end{table}

\section{TETRIS}

\subsection{Dataset statistics}
We provide statistics of image resolution (Figure~\ref{fig:resolution-plots}), number of instances per metaclass (Figure~\ref{fig:num-instances}), instance sizes (Figure~\ref{fig:instance-size}), and annotation time per image (Figure~\ref{fig:annotation-time}). Also, we provide human-related metrics for TETRIS-\textsc{people} subset only: distribution of races  (Figure~\ref{fig:races}), gender (Figure~\ref{fig:gender}), age  (Figure~\ref{fig:age}) and countries  (Figure~\ref{fig:geo}).

\subsection{Comparison to existing benchmarks}

Interactive image segmentation aims at segmenting objects of interest given an image and some sequential user input (clicks, strokes, contours). Due to a rapid evolution of hardware, modern smartphone cameras provide images of good quality and high resolution. At the same time, most interactive segmentation benchmarks -- either labeled with only boundaries SBD~\citep{sbd}, adapted~\citep{berkeley-intro,deep-object-selection,latent-diversity} from semantic segmentation datasets DAVIS, Berkeley~\citep{davis,berkeley}, or created specifically for instance segmentation GrabCut~\citep{grabcut}, -- contain images at significantly lower resolution. Benchmarking on low-resolution data may not represent the relative quality of interactive segmentation methods designed to process high-resolution images. For high-resolution images, it is crucial to keep well-defined object boundaries while editing, since accidentally smoothing them worsens the general impression and leads to less visibly plausible results~\citep{cheng2021boundary}. Accordingly, we claim most existing datasets are non-representative for assessing quality of high-resolution interactive segmentation. Besides, existing evaluation protocols do not measure boundary quality of segmented objects. A commonly used IoU measure is boundary-insensitive, thus providing incomplete evaluation in case of high-resolution images.

To provide a complete picture, we compare our dataset to the recently introduced semantic segmentation and matting datasets in Table~\ref{tbl:is-datasets}. 

Besides formal, objective, quantitative characteristics such as number of images, number of instances, image resolution etc., usability is the case. Usability is an integral characteristic, implying proper licensing, no privacy violation (all depicted people should give an articulated consent), and any other issues that may limit or forbid using a dataset.

On the contrary to the synthetic Composition-1K~\cite{xu2017deepmatting}, containing real objects inserted into background images from Pascal VOC and COCO, TETRIS comprises real data manually labeled by human annotators. Accordingly, the domain gap between TETRIS and real photos is smaller, and the results obtained for TETRIS represent in-the-wild interactive segmentation quality more adequately. 

BIG~\cite{cheng2020cascadepsp} is actually small, featuring only 50 validation and 100 test images, with one object per image. 

AIM500~\cite{jizhizi2021automatic} is also smaller than TETRIS, having only 500 samples. It also has unclear image license and apparently contains 100 photos of people for whom consent to use may not been obtained. 

AM-2k~\cite{jizhizi2021composite} contains only images of animals, and PM-10k~\cite{jizhizi2021composite} consists of human portraits. Both of them belong to a narrower domain, while TETRIS does not limit itself to a particular category but aims to cover a wide range of scenarios. AM-2k has the same licensing issues as AIM500.

\begin{table*}[t]
\fontsize{9pt}{11pt}\selectfont
\tabcolsep=2pt
\begin{tabular}{|c|c|c|l|ccc|l|ccc|l|ccc|l|ccc|l|ccc|ll}
\cline{1-3} \cline{5-7} \cline{9-11} \cline{13-15} \cline{17-19} \cline{21-23}
\multirow{3}{*}{Method}      & \multirow{3}{*}{Model} & \multirow{3}{*}{Data}                                                    &                   & \multicolumn{3}{c|}{GrabCut}                    &  & \multicolumn{3}{c|}{Berkeley}                   &  & \multicolumn{3}{c|}{DAVIS}                      &  & \multicolumn{3}{c|}{COCO-MVal}                  &  & \multicolumn{3}{c|}{TETRIS (ours)}              &                      &                      \\ \cline{5-7} \cline{9-11} \cline{13-15} \cline{17-19} \cline{21-23}
                             &                        &                                                                          &                   & \multicolumn{3}{c|}{BIoU (AuC@10)}              &  & \multicolumn{3}{c|}{BIoU (AuC@10)}              &  & \multicolumn{3}{c|}{BIoU (AuC@10)}              &  & \multicolumn{3}{c|}{BIoU (AuC@10)}              &  & \multicolumn{3}{c|}{BIoU (AuC@10)}              &                      &                      \\
                             &                        &                                                                          &                   & Min$\uparrow$            & Max$\uparrow$            & D$\downarrow$             &  & Min$\uparrow$            & Max$\uparrow$            & D$\downarrow$             &  & Min$\uparrow$            & Max$\uparrow$            & D$\downarrow$             &  & Min$\uparrow$            & Max$\uparrow$            & D$\downarrow$             &  & Min$\uparrow$            & Max$\uparrow$            & D$\downarrow$             &                      &                      \\ \cline{1-3} \cline{5-7} \cline{9-11} \cline{13-15} \cline{17-19} \cline{21-23}
MobileSAM                    & ViT-Tiny               & SA-1B                                                                    &                   & 84.67          & 87.54          & 2.87          &  & 80.41          & 83.42          & 3.01          &  & 75.73          & 80.11          & 4.38          &  & 68.34          & 73.49          & 5.15          &  & 75.66          & 81.64          & 5.98          &                      &                      \\ \cline{1-3} \cline{5-7} \cline{9-11} \cline{13-15} \cline{17-19} \cline{21-23}
\multirow{3}{*}{SAM}         & ViT-B                  & \multirow{3}{*}{SA-1B}                                                   &                   & 85.03          & 88.63          & 3.60          &  & 81.84          & 85.31          & 3.47          &  & 75.95          & 82.05          & 6.09          &  & 70.93          & 75.79          & 4.86          &  & 75.82          & 83.69          & 7.87          &                      &                      \\ \cline{2-2}
                             & ViT-L                  &                                                                          &                   & 86.82          & 88.72          & \textbf{1.90} &  & 84.39          & 86.27          & \textbf{1.88} &  & 77.25          & 82.69          & 5.44          &  & 72.76          & 76.02          & \textbf{3.26} &  & 79.52          & 85.39          & 5.88          &                      &                      \\ \cline{2-2}
                             & ViT-H                  &                                                                          &                   & 86.01          & 88.53          & 2.53          &  & 83.30          & 85.99          & 2.70          &  & 77.20          & 82.32          & 5.12          &  & 70.84          & 74.30          & \underline{3.46}    &  & 79.65          & 85.13          & 5.48          &                      &                      \\ \cline{1-3} \cline{5-7} \cline{9-11} \cline{13-15} \cline{17-19} \cline{21-23}
\multirow{3}{*}{SAM-HQ}      & ViT-B                  & \multirow{3}{*}{\begin{tabular}[c]{@{}c@{}}SA-\\ 1B\\ +44K\end{tabular}} &                   & 83.59          & 89.30          & 5.71          &  & 82.36          & 87.31          & 4.95          &  & 75.08          & 83.37          & 8.29          &  & 70.52          & 77.13          & 6.61          &  & 70.59          & 86.65          & 16.06         &                      &                      \\ \cline{2-2}
                             & ViT-L                  &                                                                          &                   & \textbf{87.39} & 90.22          & 2.83          &  & 85.38          & 88.16          & 2.78          &  & 75.55          & 83.93          & 8.37          &  & 72.87          & 78.13          & 5.26          &  & 78.35          & 88.25          & 9.90          &                      &                      \\ \cline{2-2}
                             & ViT-H                  &                                                                          &                   & 86.60          & \underline{90.50}    & 3.91          &  & 85.52          & 88.34          & 2.83          &  & 74.14          & 84.03          & 9.88          &  & 72.23          & 78.42          & 6.19          &  & 78.58          & 88.25          & 9.67          &                      &                      \\ \cline{1-3} \cline{5-7} \cline{9-11} \cline{13-15} \cline{17-19} \cline{21-23}
\multirow{2}{*}{CDNet}       & RN34                   & C+L                                                                      &                   & 80.74          & 88.26          & 7.52          &  & 79.87          & 86.05          & 6.18          &  & 75.68          & 80.63          & 4.96          &  & 71.22          & 79.82          & 8.61          &  & 77.72          & 84.39          & 6.67          &                      &                      \\ \cline{2-3}
                             & RN34                   & SBD                                                                      &                   & 72.76          & 81.52          & 8.75          &  & 74.71          & 80.55          & 5.84          &  & 71.77          & 76.98          & 5.22          &  & 60.76          & 71.88          & 11.12         &  & 69.74          & 77.66          & 7.92          & \multicolumn{1}{c}{} &                      \\ \cline{1-3} \cline{5-7} \cline{9-11} \cline{13-15} \cline{17-19} \cline{21-23}
GPCIS                        & RN50                   & C+L                                                                      &                   & 79.67          & 86.32          & 6.65          &  & 77.65          & 85.22          & 7.57          &  & 75.30          & 80.58          & 5.28          &  & 66.92          & 81.75          & 14.83         &  & 72.18          & 83.10          & 10.92         &                      &                      \\ \cline{1-3} \cline{5-7} \cline{9-11} \cline{13-15} \cline{17-19} \cline{21-23}
\multirow{5}{*}{RITM}        & HR18s-IT               & \multirow{4}{*}{C+L}                                                     &                   & 79.80          & 86.11          & 6.31          &  & 81.63          & 84.87          & 3.24          &  & 68.58          & 78.58          & 10.00         &  & 75.95          & 82.73          & 6.78          &  & 76.98          & 82.09          & 5.11          &                      &                      \\ \cline{2-2}
                             & HR18                   &                                                                          &                   & 80.07          & 83.93          & 3.86          &  & 80.60          & 83.48          & 2.88          &  & 74.21          & 78.63          & 4.42          &  & 75.23          & 80.07          & 4.84          &  & 75.55          & 79.30          & 3.75          & \multicolumn{1}{c}{} & \multicolumn{1}{c}{} \\ \cline{2-2}
                             & HR18-IT                &                                                                          &                   & 83.54          & 85.81          & 2.27          &  & 83.76          & 86.36          & 2.60          &  & 72.16          & 79.99          & 7.83          &  & 78.35          & 84.25          & 5.90          &  & 79.34          & 83.11          & 3.77          & \multicolumn{1}{c}{} & \multicolumn{1}{c}{} \\ \cline{2-2}
                             & HR32-IT                &                                                                          &                   & 82.81          & 86.72          & 3.91          &  & 83.31          & 87.03          & 3.71          &  & 71.31          & 81.10          & 9.79          &  & 77.23          & 84.54          & 7.31          &  & 78.72          & 83.92          & 5.20          & \multicolumn{1}{c}{} & \multicolumn{1}{c}{} \\ \cline{2-3} \cline{5-7} \cline{9-11} \cline{13-15} \cline{17-19} \cline{21-23}
                             & HR18-IT                & SBD                                                                      &                   & 77.77          & 82.00          & 4.23          &  & 76.74          & 80.92          & 4.18          &  & 72.32          & 77.02          & 4.71          &  & 67.73          & 77.06          & 9.32          &  & 70.00          & 75.75          & 5.75          & \multicolumn{1}{c}{} &                      \\ \cline{1-3} \cline{5-7} \cline{9-11} \cline{13-15} \cline{17-19} \cline{21-23}
\multirow{7}{*}{SimpleClick} & ViT-B                  & \multirow{3}{*}{C+L}                                                     &                   & 84.99          & 88.49          & 3.50          &  & \underline{86.25}          & 88.42          & 2.17          &  & 79.59          & 83.39          & 3.80          &  & 81.14          & 85.57          & 4.43          &  & 83.20          & 87.23          & 4.02          &                      &                      \\ \cline{2-2}
                             & ViT-L                  &                                                                          &                   & 86.54          & 89.26          & 2.73          &  & 85.68          & \underline{89.31}          & 3.63          &  & \underline{81.54}    & \underline{84.29}    & \underline{2.75}    &  & \underline{82.85}    & \underline{86.80}    & 3.95          &  & 84.76          & 88.32          & \underline{3.56}    & \multicolumn{1}{c}{} & \multicolumn{1}{c}{} \\ \cline{2-2}
                             & ViT-H                  &                                                                          &                   & 85.77          & 89.38          & 3.61          &  & 86.17          & 89.26          & 3.09          &  & 80.16          & 84.11          & 3.95          &  & \textbf{83.28} & \textbf{87.03} & 3.75          &  & \underline{85.37}    & \underline{88.62}    & \textbf{3.25} & \multicolumn{1}{c}{} & \multicolumn{1}{c}{} \\ \cline{2-3} \cline{5-7} \cline{9-11} \cline{13-15} \cline{17-19} \cline{21-23}
                             & ViT-XT                 & \multirow{4}{*}{SBD}                                                     & \multirow{4}{*}{} & 79.80          & 82.79          & 2.99          &  & 75.66          & 80.17          & 4.51          &  & 66.81          & 75.46          & 8.65          &  & 66.50          & 76.97          & 10.47         &  & 67.21          & 76.02          & 8.81          &                      &                      \\ \cline{2-2}
                             & ViT-B                  &                                                                          &                   & 83.62          & 86.50          & 2.88          &  & 83.05          & 85.06          & 2.01    &  & 78.86          & 81.62          & 2.76          &  & 72.08          & 79.97          & 7.90          &  & 76.36          & 80.88          & 4.52          & \multicolumn{1}{c}{} & \multicolumn{1}{c}{} \\ \cline{2-2}
                             & ViT-L                  &                                                                          &                   & 83.33          & 86.38          & 3.05          &  & 82.98          & 85.15          & 2.17          &  & 79.41          & 81.67          & \textbf{2.27} &  & 74.66          & 81.75          & 7.09          &  & 77.90          & 81.75          & 3.85          & \multicolumn{1}{c}{} & \multicolumn{1}{c}{} \\ \cline{2-2}
                             & ViT-H                  &                                                                          &                   & 84.57          & 86.74          & \underline{2.17}    &  & 83.41          & 85.35          & \underline{1.95}          &  & 78.88          & 81.72          & 2.84          &  & 74.54          & 81.95          & 7.41          &  & 77.76          & 81.56          & 3.81          & \multicolumn{1}{c}{} & \multicolumn{1}{c}{} \\ \cline{1-3} \cline{5-7} \cline{9-11} \cline{13-15} \cline{17-19} \cline{21-23}
CFR-ICL                      & ViT-H                  & C+L                                                                      &                   & \underline{87.14}    & \textbf{90.97} & 3.83          &  & \textbf{87.53} & \textbf{91.05} & 3.51          &  & \textbf{81.61} & \textbf{85.85} & 4.24          &  & 81.62          & 86.57          & 4.94          &  & \textbf{85.91} & \textbf{90.71} & 4.80          &                      &                      \\ \cline{1-3} \cline{5-7} \cline{9-11} \cline{13-15} \cline{17-19} \cline{21-23}
\end{tabular}
\caption{The BIoU quality and robustness scores of different models, measured on the standard datasets and our novel TETRIS dataset. The best results are \textbf{bold}, the second best are \underline{underlined}.}
\label{tbl:base-d-og}
\end{table*}

PM-10k is not open-sourced due to privacy issues. However, the version of this dataset undergone the privacy-preserving face blurring is published as P3M-10k. However, there is still a concern about privacy, since a person can be identified even if their face is blurred.

\begin{figure*}[h]
    \centering
    \includegraphics[width=0.975\linewidth]{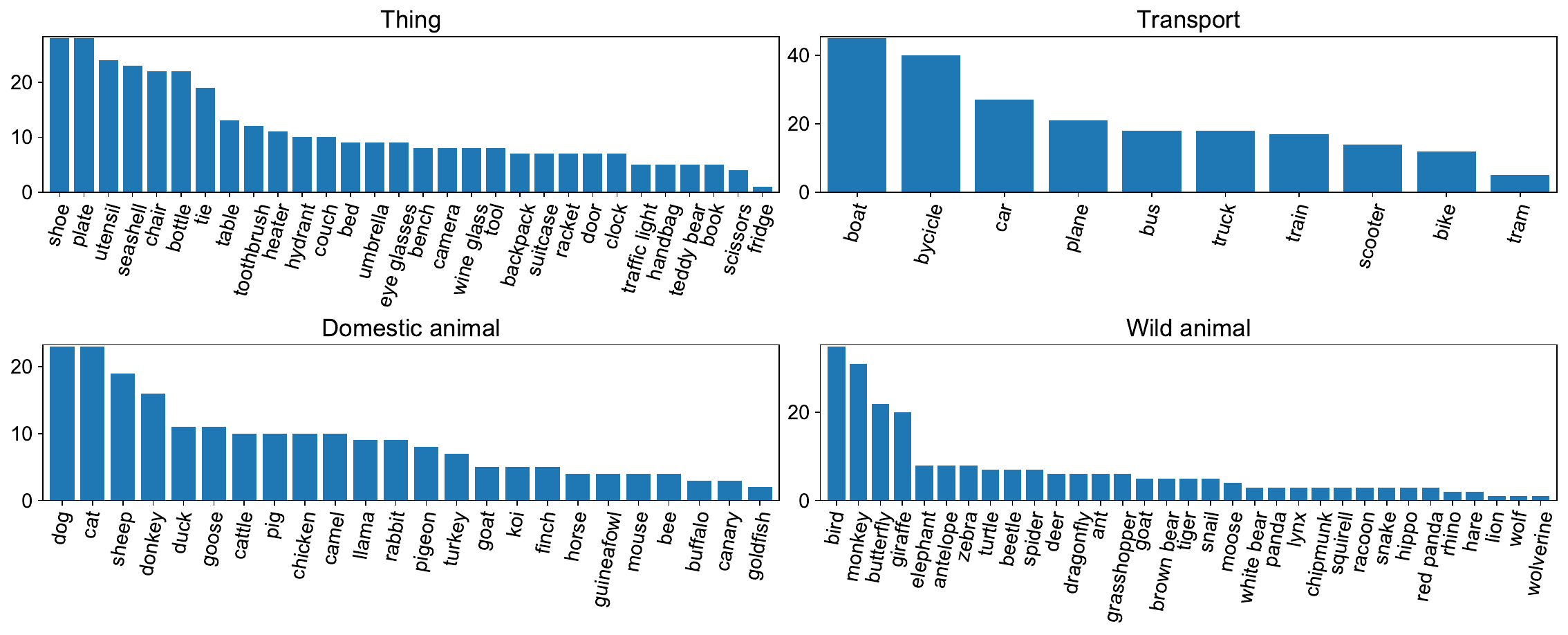}
    \captionof{figure}{
        Number of instances per class for four metaclasses.
    }
    \label{fig:istats}
\end{figure*}

\subsection{Object Classes Selection}
\label{three_part}

There exist two approaches to object classes selection: informally, aiming either for quantity or quality. The “quantity-focused” approach implies using a large-scale hierarchical lexical database, which is then coarsely filtered using heuristic rules in an automatic manner. Such an approach is the most suitable for acquisition of universal voluminous image collections such as ImageNet~\citep{imagenet} and OpenImages~\citep{open-images} with each class being represented with numerous images. However, the collection might contain images of poor quality, explicit content, ambiguous examples, duplicates, and other types of noisy, inappropriate, or misleading data.

The alternative “quality-focused” approach is to manually select classes of interest based on task-specific requirements (e.g. COCO~\citep{coco,coco-stuff}). It provides fewer images in total and might result in long-tail class distribution, with some classes being over-represented and many others having only a few images. Collecting our own benchmark, we follow the “quality-focused” approach, yet aiming to maintain the well-balanced class distribution.

Specifically, we choose object classes according to task-specific requirements. We also rely on previous works and include classes from PASCAL VOC 2012~\citep{pascal-voc-2012} and some other common classes from COCO~\citep{coco}. Overall, we consider 9 metaclasses: \textit{transport, architecture, monument, thing, wild animal, domestic animal, plant, food, people}. We show the distribution of instances of fine-grained classes for \textit{thing, transport, domestic animal, wild animal} metaclasses in Figure~\ref{fig:stats}. Overall, there are 105 object classes in the dataset (Figure~\ref{fig:istats}). According to the histograms, some classes are represented with a larger number of images than the others; however, the class distribution is not a long-tail one.

\subsection{Data Labeling}

\paragraph{Annotators}
\label{ssec:annotators}

We contracted a company specializing in data markup for labeling our data. The company hired 10 annotators to perform the task. Each annotator was paid 200\% of the minimal wage. Total compensation depended on the time spent on annotation.

\paragraph{Contractor Specification}
\label{ssec:annotator-guide}

We provided our contractor with the following Specification:

\begin{enumerate}
    \item The annotator labels the received images with object classes and transfers back the following data:
    \begin{itemize}
        \item instance segmentation masks as PNG images of the same resolution as original RGB images; each object in a mask should have a unique color;
        \item a file containing measurements of the annotation time per image.
    \end{itemize}
    
    \item On each image, at least one object of interest should be labeled. Typically, such objects occupy the large area of an image and are located in the vicinity of the image center. Every object in an image is segmented independently: if there are several objects, the image gets segmented for several times by the same annotator. 
    
    \item The object mask is defined as a polygon following the contour of this object, assigned with an object class. These polygons are created via a CVAT labeling tool~\cite{cvat}.
    
    \item Each polygon is classified into three types:
    \begin{itemize}
        \item \textit{foreground} (object of interest, Figure 4);
        \item \textit{background} (the remaining image area that does not belong to the object of interest, Figure 5);
        \item \textit{uncertain} (regions around the foreground-background boundaries, or the boundaries between objects. Labeling such regions (Figure 6) might be challenging, especially when objects have fur, feathers, hair, or leaves).
    \end{itemize}
\end{enumerate}

    \begin{table}[h!]
            \setlength{\tabcolsep}{3pt}{
            \resizebox{1.0\linewidth}{!}{
            \begin{tabular}{ccc}
                \includegraphics[height=100pt]{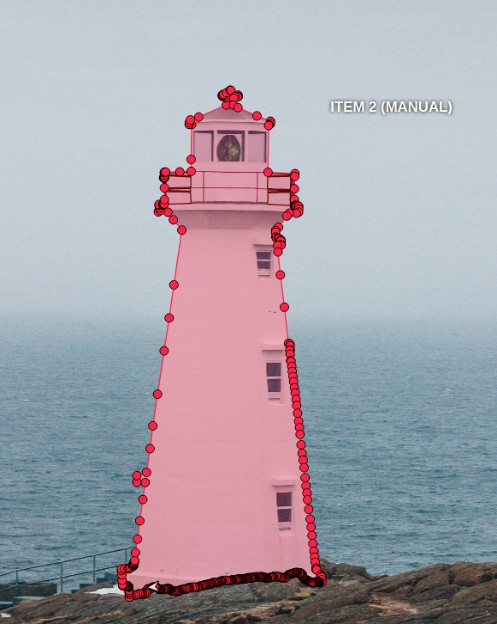} &
                \includegraphics[height=100pt]{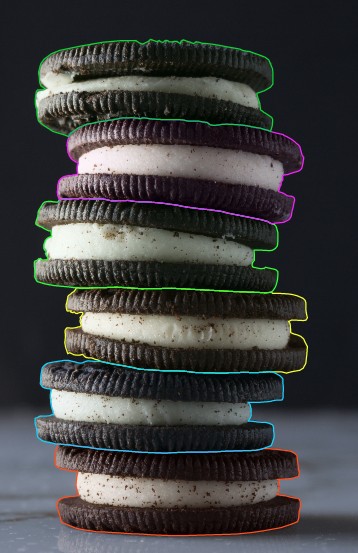} &
                \includegraphics[height=100pt]{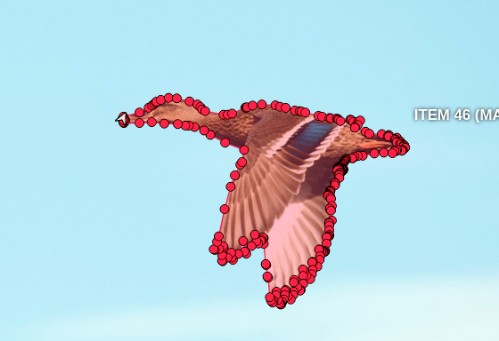} \\
            \end{tabular}
            }
            \captionof{figure}{Examples of desired foreground regions to be annotated. Red masks highlight the object of interest, red points depict polygon vertices.}
            }
    \end{table}
    \begin{table}[h!]
            \setlength{\tabcolsep}{3pt}{
            \resizebox{1.0\linewidth}{!}{
            \begin{tabular}{ccc}
                \includegraphics[height=70pt]{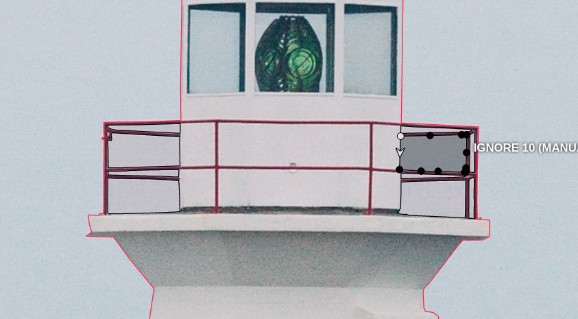} &
                \includegraphics[height=70pt]{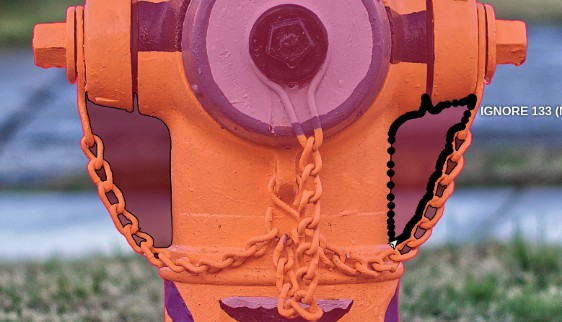} &
                \includegraphics[height=70pt]{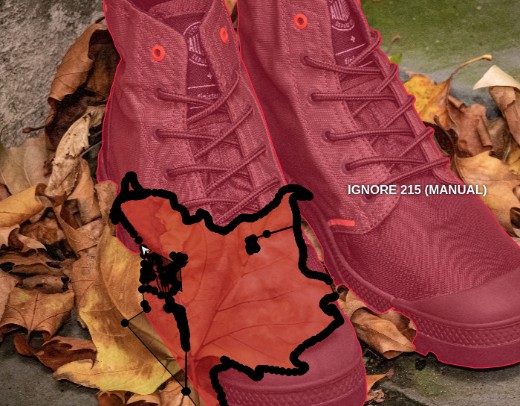} \\
            \end{tabular}
            }
            \label{fig:back}
            \captionof{figure}{Examples of desired background regions to be annotated. Orange masks define the background regions with black points marking polygon vertices.}
            }
    \end{table}

    \begin{table}[h!]
            \setlength{\tabcolsep}{3pt}{
            \resizebox{1.0\linewidth}{!}{
            \begin{tabular}{ccc}
                \includegraphics[height=100pt]{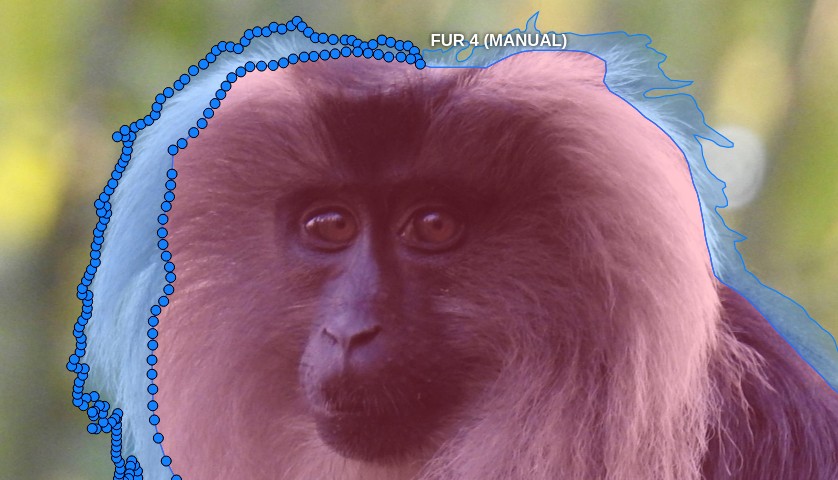} &
                \includegraphics[height=100pt]{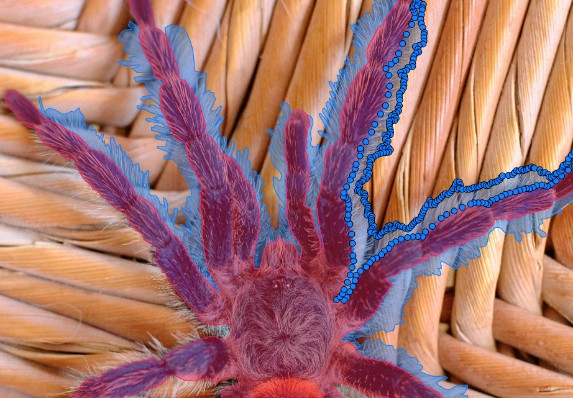} &
                \includegraphics[height=100pt]{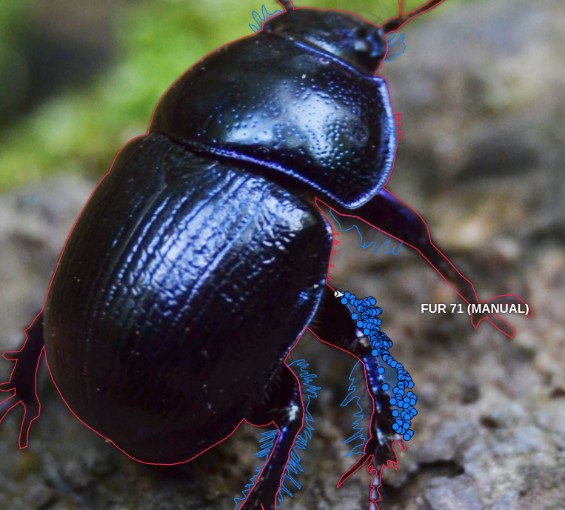} \\
            \end{tabular}
            }
            \label{fig:uncert}
            \captionof{figure}{Examples of desired uncertain regions to be annotated. Uncertain regions are blue with blue points depicting polygon vertices.}
            }
    \end{table}
    
\paragraph{Refining Annotated Masks via Matting}
\label{ssec:matting}

We receive object masks in the form of polygons classified either as foreground, background, or uncertain. Then, we refine object boundaries via matting. Specifically, we apply the current state-of-the-art matting method~\cite{matteformer} per each segmented object and obtain an alpha map. If an image depicts several objects, we merge corresponding alpha maps, assigning each pixel to the object with a larger alpha value. Next, we threshold merged alpha maps, that gives a binary segmentation mask. However, matting provides non-satisfactory results in some cases (5\% of all processed images), so we manually correct the erroneous regions to obtain final masks. Matting was used primarily for labeling domestic and wild animals, which are covered with fur. Moreover, it was for labelling a few images of cacti (“plant” metaclass) and peoples as well.

\begin{figure}[h]
    \centering
    \includegraphics[width=1.0\linewidth]{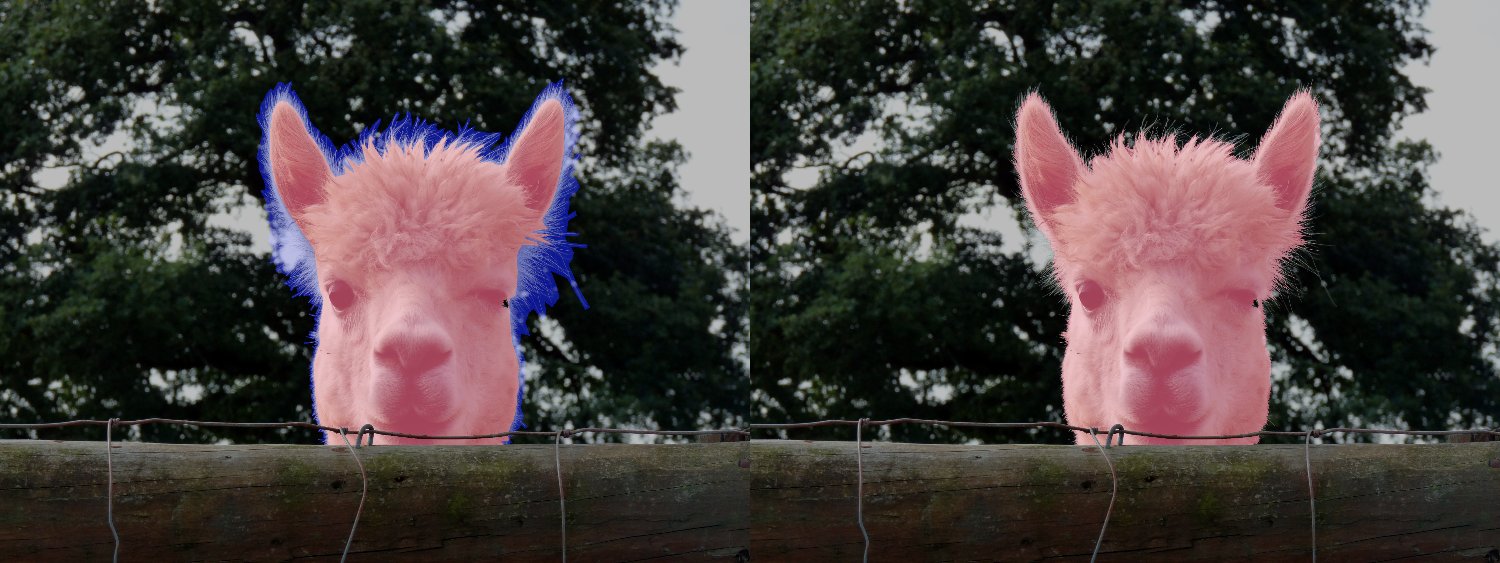}
    \includegraphics[width=1.0\linewidth]{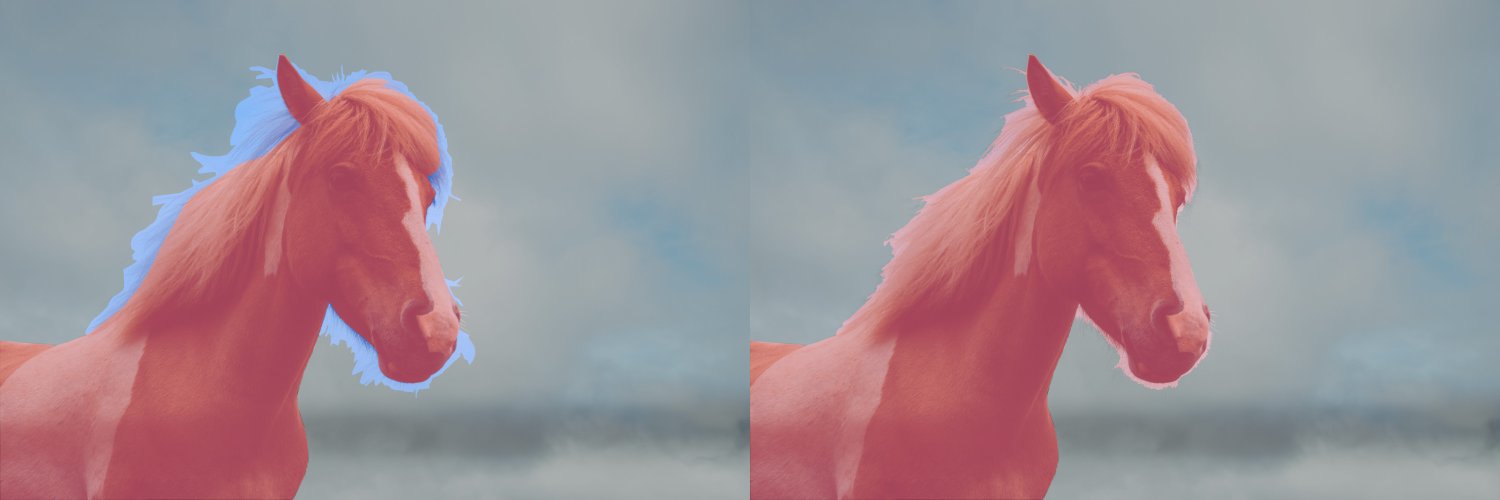}
    \caption{Post-processing manually labeled masks for images containing a single object via matting. The image regions classified as foreground are colored red, uncertain regions are blue. In each row, the left image depicts the mask provided by a human annotator, the right image visualizes the result of matting.}
    \label{fig:matting-examples}
\end{figure}

We compare masks before and after matting in Figure~\ref{fig:matting-examples}. As can be observed from the visualized examples, matting provides decent results, yet not always being pixel-wise accurate. 

\begin{figure}[ht]
    \centering
    \includegraphics[width=1.0\linewidth]{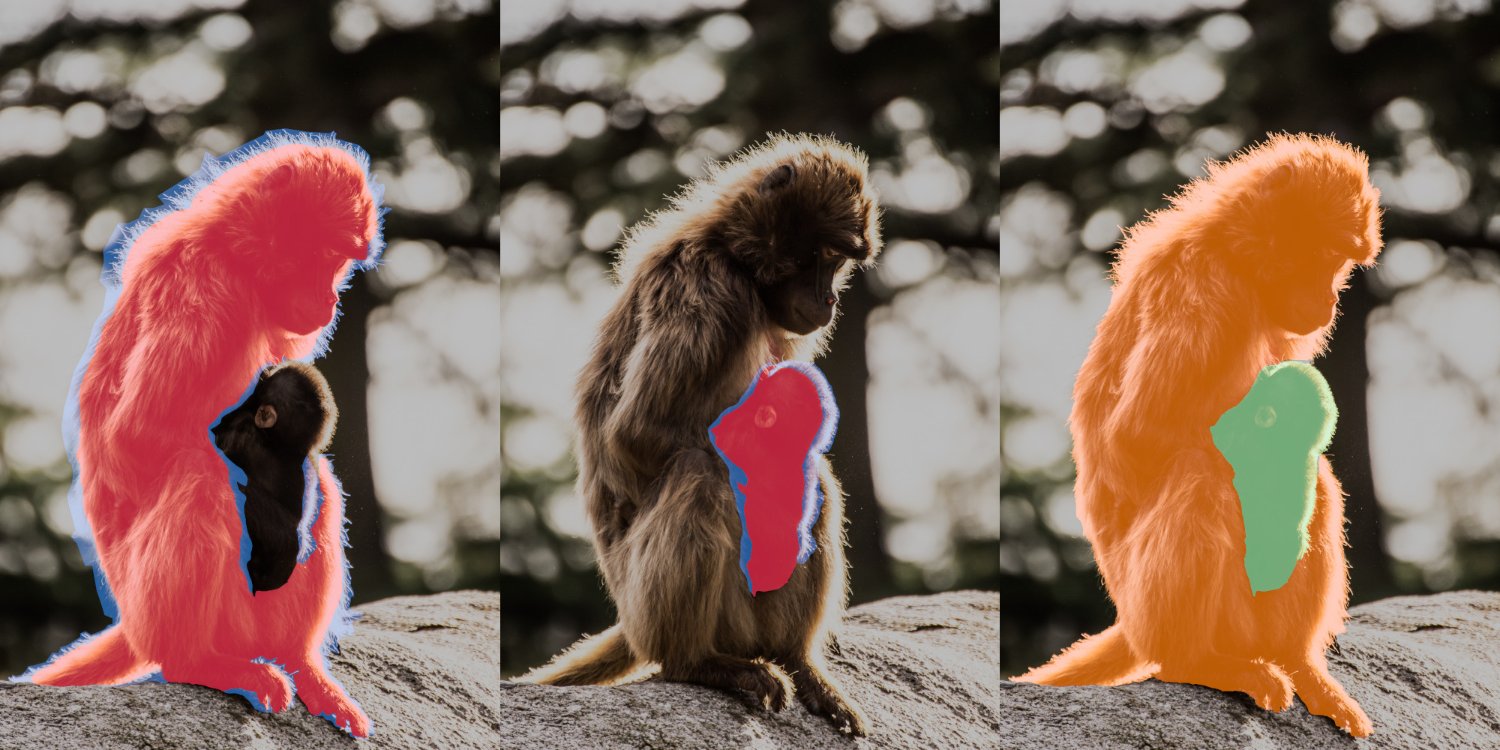}
    \includegraphics[width=1.0\linewidth]{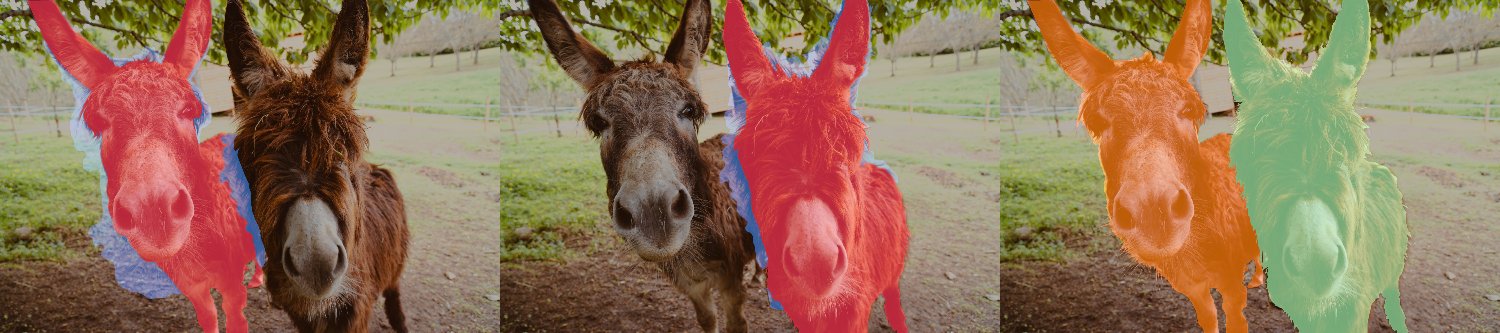}
    \caption{Post-processing manually labeled masks for images of several objects via matting. The image regions classified as foreground are colored red, uncertain regions are blue. In each row, all images, but the right one depict the mask provided by a human annotator for each object, the right image visualizes the result of matting.}
    \label{fig:matting-several-objects}
\end{figure}

\begin{figure*}
    \centering
    \includegraphics[width=\linewidth]{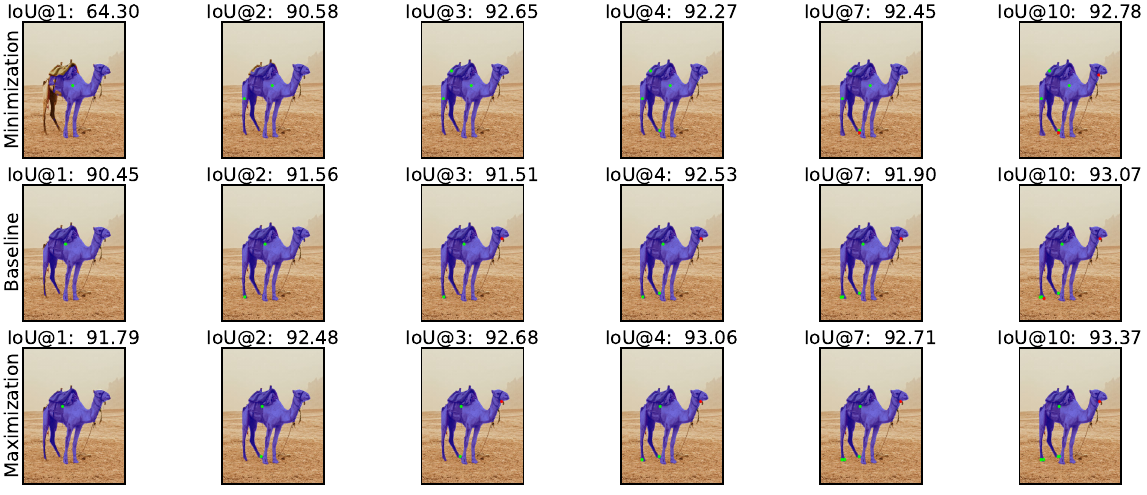}
    \includegraphics[width=\linewidth]{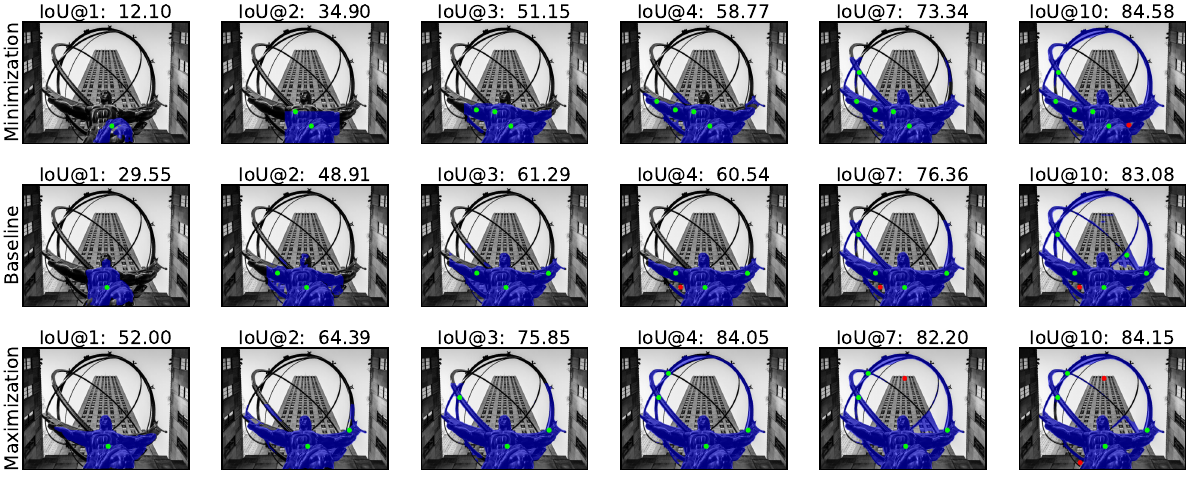}
    \includegraphics[width=\linewidth]{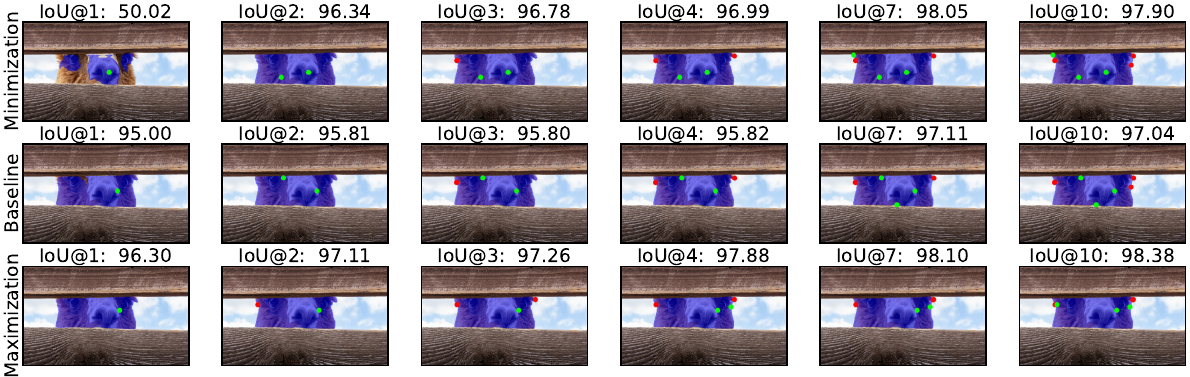}
    \captionof{figure}{
        Visualization of \textit{minimization, baseline, maximization} clicks trajectories for RITM-HRNet18-IT~\cite{ritm} with predicted masks (blue overlap), positive (green) and negative (red) clicks.
    }
    \label{fig:traj1}
\end{figure*}

\begin{figure*}
    \centering
    \includegraphics[width=\linewidth]{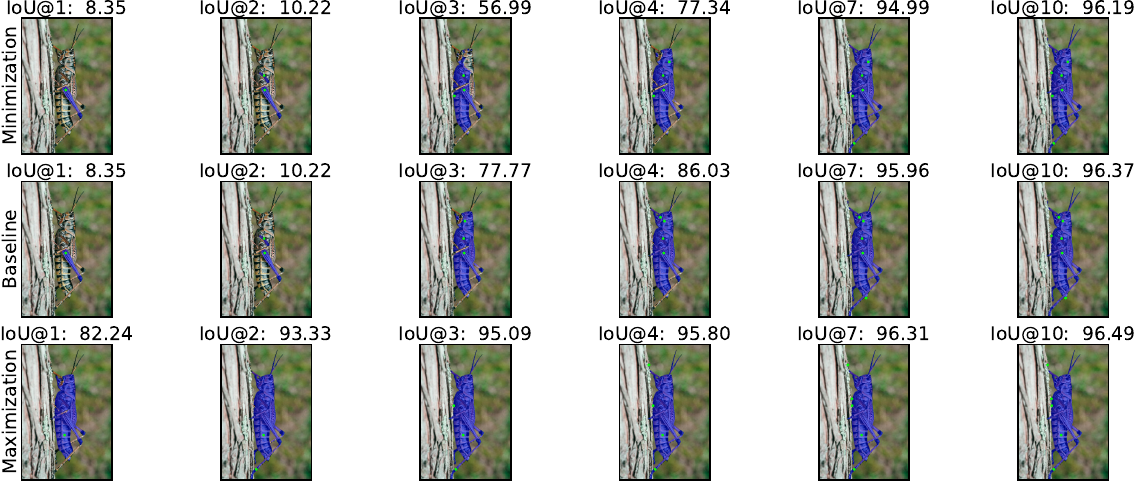}
    \includegraphics[width=\linewidth]{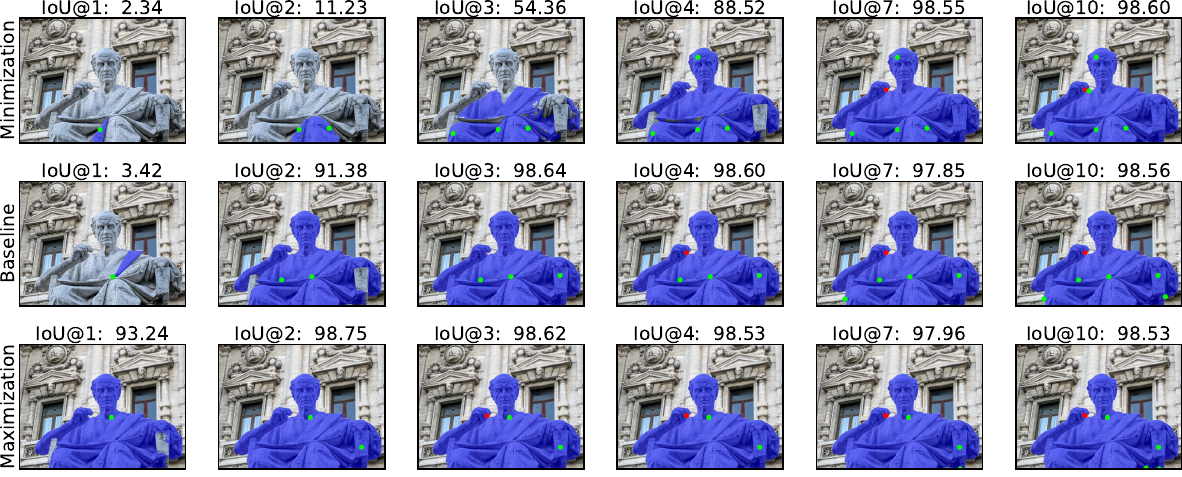}
    \includegraphics[width=\linewidth]{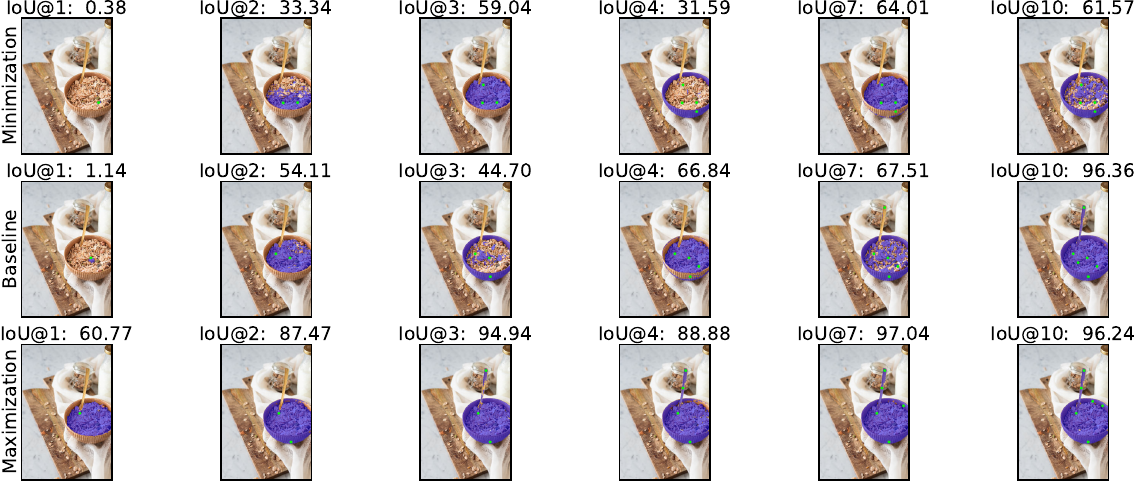}
    \captionof{figure}{
        Visualization of \textit{minimization, baseline, maximization} clicks trajectories for SAM-ViT-H~\cite{kirillov2023segment} with predicted masks (blue overlap), positive (green) and negative (red) clicks.
    }
    \label{fig:traj2}
\end{figure*}

\begin{figure*}
    \centering
    \includegraphics[width=1.0\textwidth]{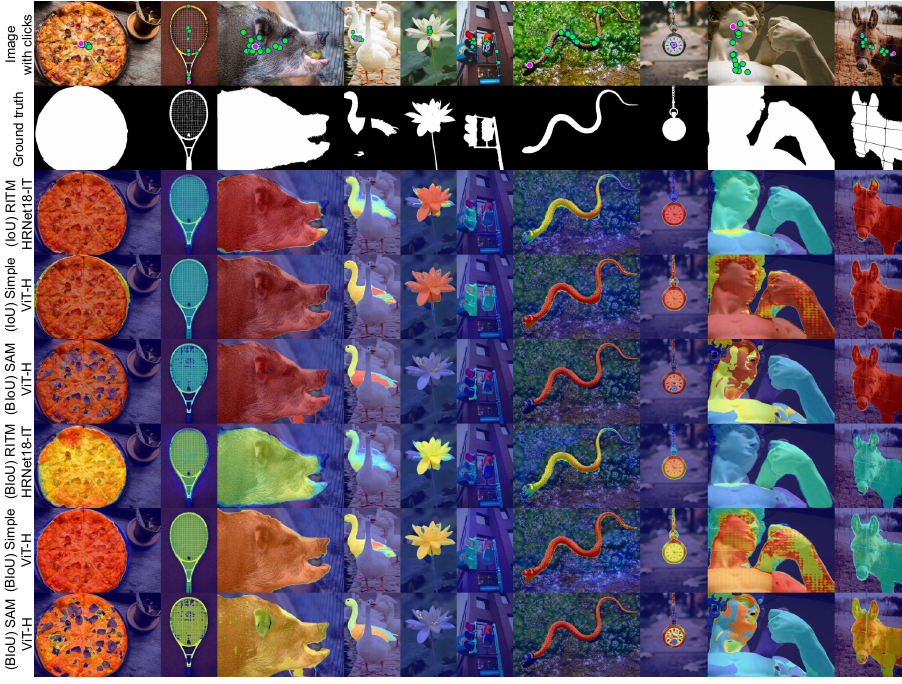}
    \caption{Top rows: images with real-user clicks (green) and clicks, obtained from baseline strategy (magenta); ground-truth binary masks. Three next rows: IoU scores of RITM, SimpleClick, SAM, calculated extensively on a grid for every possible integer click position. Three next rows: corresponding Boundary-IoU (BIoU) scores of RITM, SimpleClick, SAM. Higher scores are visualized with warmer colors.}
    \label{fig:supplem_user_study1}
\end{figure*}

\begin{figure*}
    \includegraphics[width=1.0\textwidth]{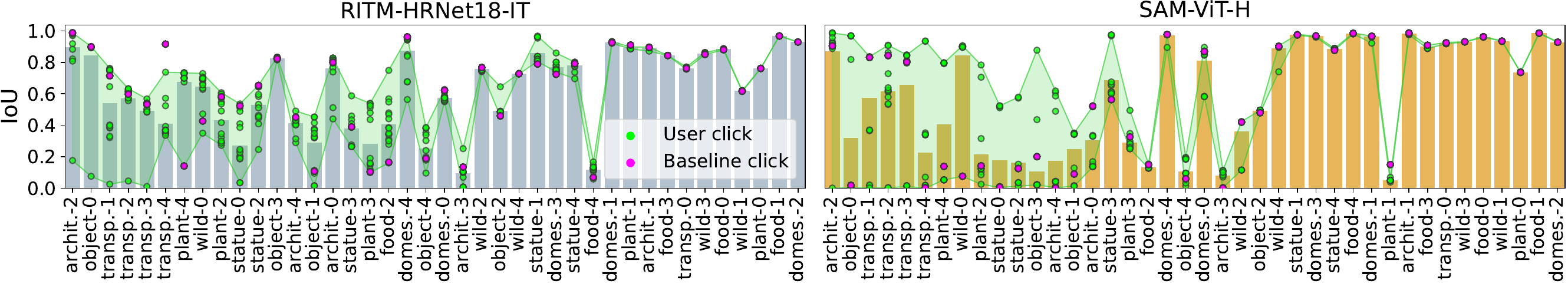}
    \caption{An IoU spread (a difference between a maximum and minimum IoU of user clicks) between predicted and ground truth masks in the first real user interaction round. Green points represent user clicks, magenta points depict the clicks generated with the baseline strategy. Columns are sorted by an average spread.}
    \label{fig:user-study-std11}
\end{figure*}

\begin{figure*}
    \centering
    \includegraphics[width=1.0\textwidth]{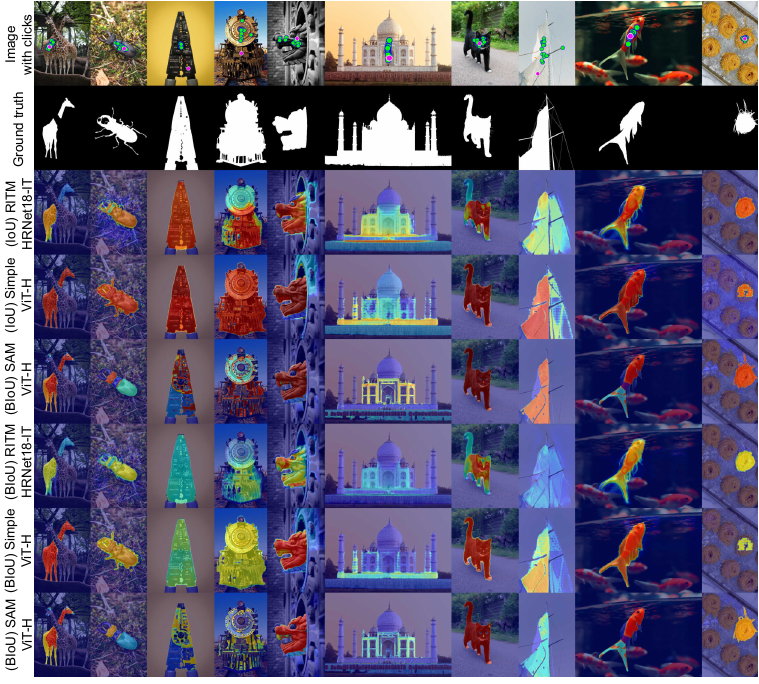}
    \caption{Top rows: images with real-user clicks (green) and clicks, obtained from baseline strategy (magenta); ground-truth binary masks. Three next rows: IoU scores of RITM, SimpleClick, SAM, calculated extensively on a grid for every possible integer click position. Three next rows: corresponding Boundary-IoU (BIoU) scores of RITM, SimpleClick, SAM. Higher scores are visualized with warmer colors.}
    \label{fig:supplem_user_study2}
    \includegraphics[width=1.0\textwidth]{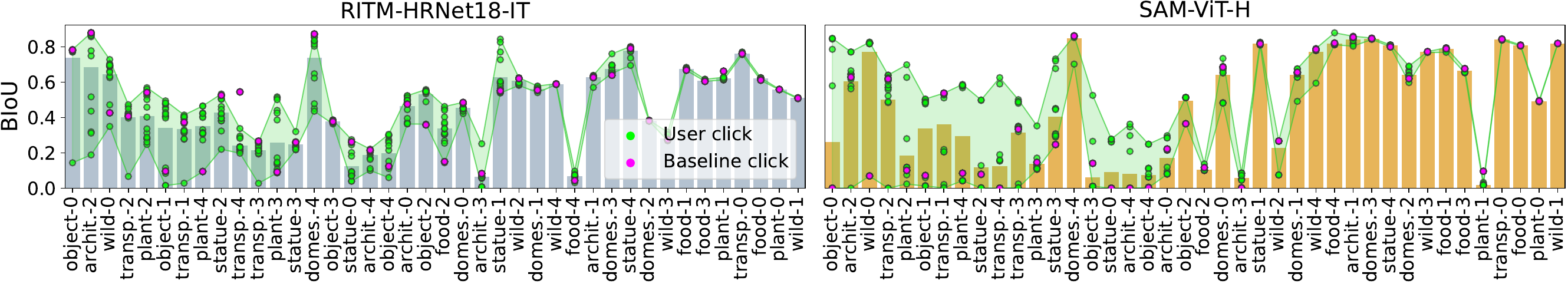}
    \caption{A BIoU spread (a difference between a maximum and minimum BIoU of user clicks) between predicted and ground truth masks in the first real user interaction round. Green points represent user clicks, magenta points depict the clicks generated with the baseline strategy. Columns are sorted by an average spread.}
    \label{fig:user-study-std12}
\end{figure*}

\begin{figure*}
    \centering
    \includegraphics[width=1.0\textwidth]{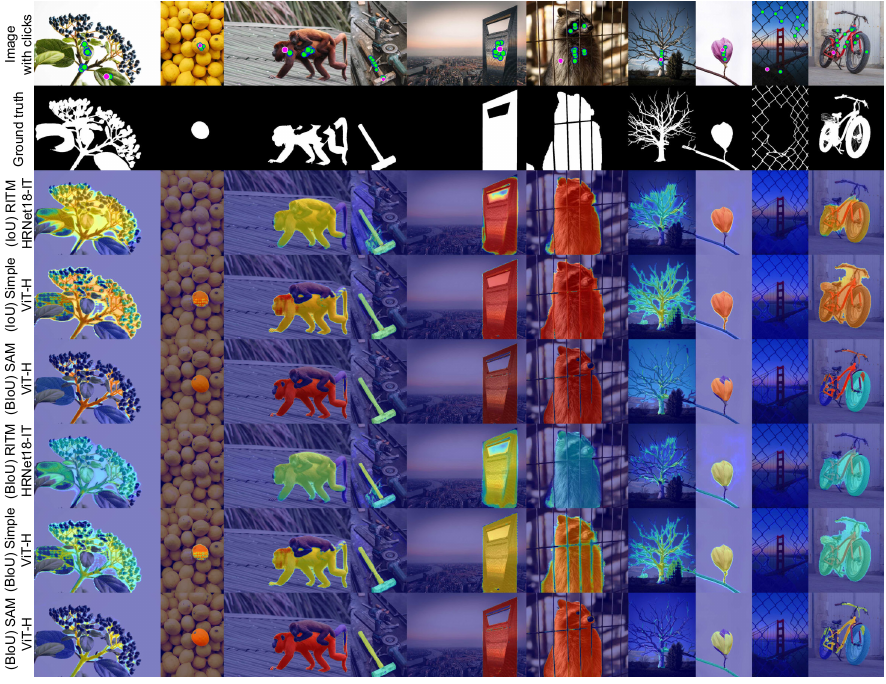}
    \caption{Top rows: images with real-user clicks (green) and clicks, obtained from baseline strategy (magenta); ground-truth binary masks. Three next rows: IoU scores of RITM, SimpleClick, SAM, calculated extensively on a grid for every possible integer click position. Three next rows: corresponding Boundary-IoU (BIoU) scores of RITM, SimpleClick, SAM. Higher scores are visualized with warmer colors.}
    \label{fig:supplem_user_study3}
\end{figure*}

\begin{figure*}    
    \centering
    \includegraphics[width=1.0\textwidth]{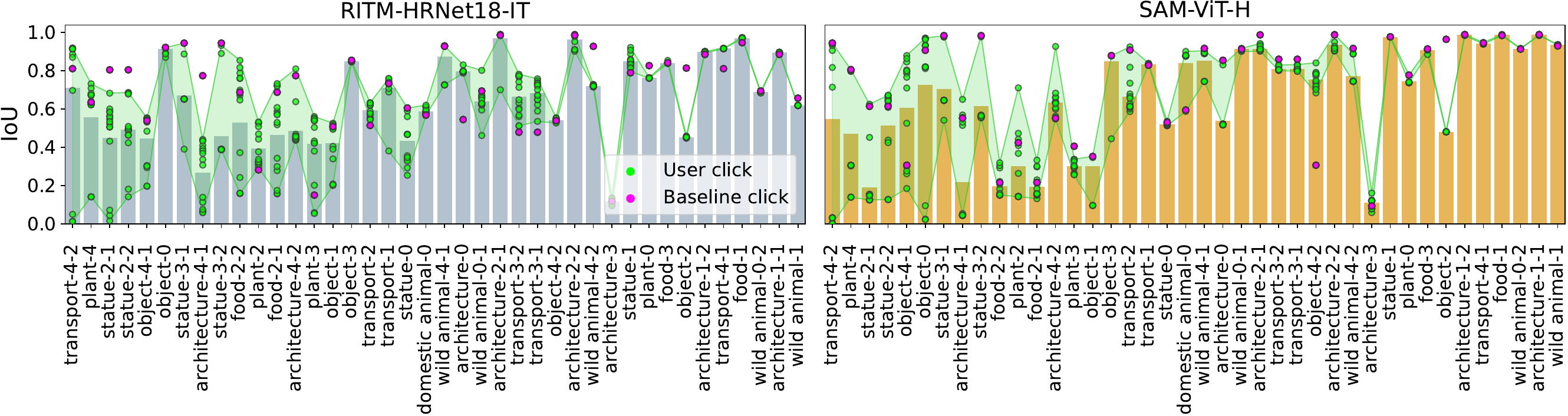}
    \caption{An IoU spread (a difference between a maximum and minimum IoU of user clicks) between predicted and ground truth masks in the second positive real user interaction round. Green points represent user clicks, magenta points depict the clicks generated with the baseline strategy. Columns are sorted by an average spread.}
    \label{fig:user-study-std21}
\end{figure*}

\begin{figure*}
    \centering
    \includegraphics[width=1.0\textwidth]{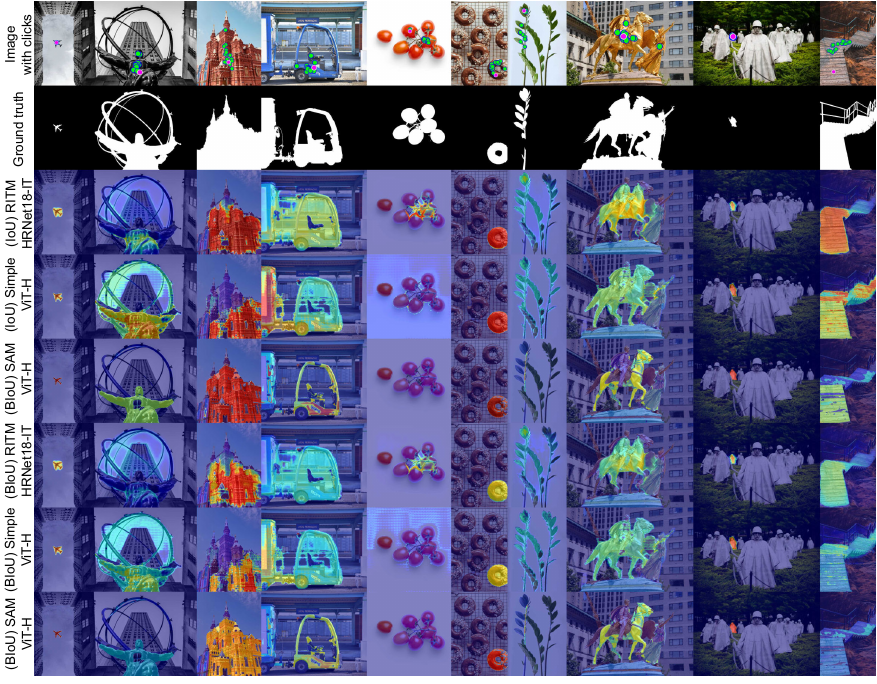}
    \caption{Top rows: images with real-user clicks (green) and clicks, obtained from baseline strategy (magenta); ground-truth binary masks. Three next rows: IoU scores of RITM, SimpleClick, SAM, calculated extensively on a grid for every possible integer click position. Three next rows: corresponding Boundary-IoU (BIoU) scores of RITM, SimpleClick, SAM. Higher scores are visualized with warmer colors.}
    \label{fig:supplem_user_study4}
 
    \centering
    \includegraphics[width=1.0\textwidth]{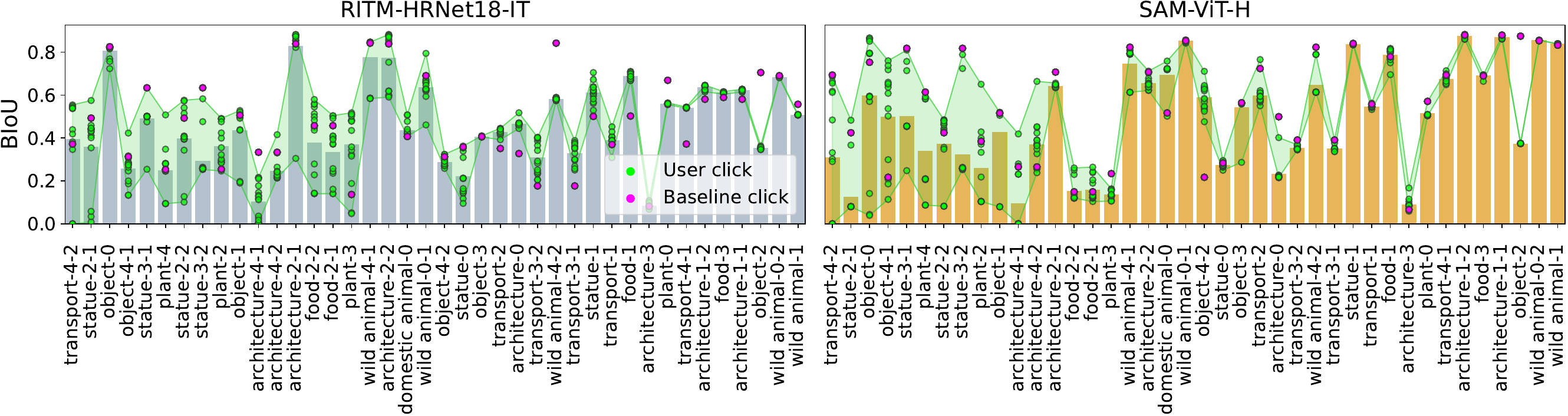}
    \caption{A BIoU spread (a difference between a maximum and minimum BIoU of user clicks) between predicted and ground truth masks in the second positive real user interaction round. Green points represent user clicks, magenta points depict the clicks generated with the baseline strategy. Columns are sorted by an average spread.}
    \label{fig:user-study-std22}
\end{figure*}

\begin{figure*}
    \centering
    \includegraphics[width=1.0\textwidth]{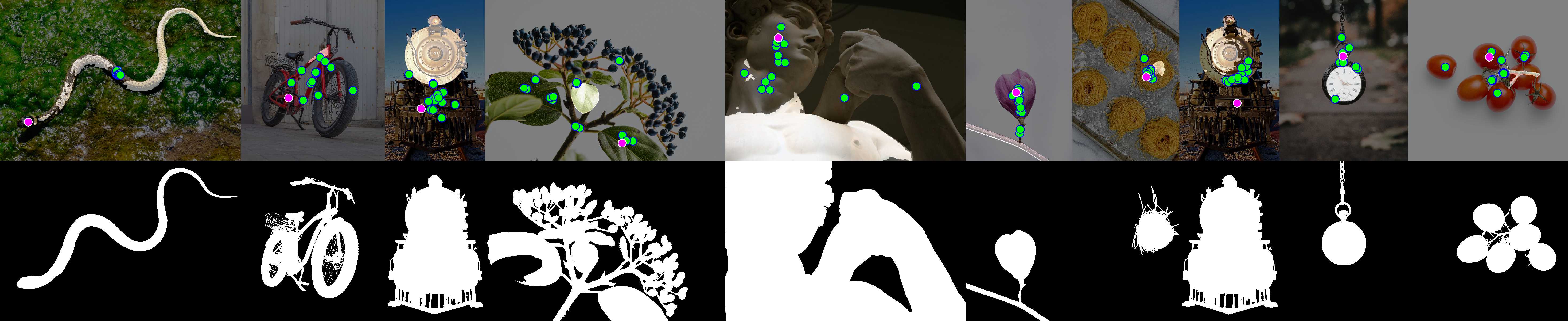}
    \includegraphics[width=1.0\textwidth]{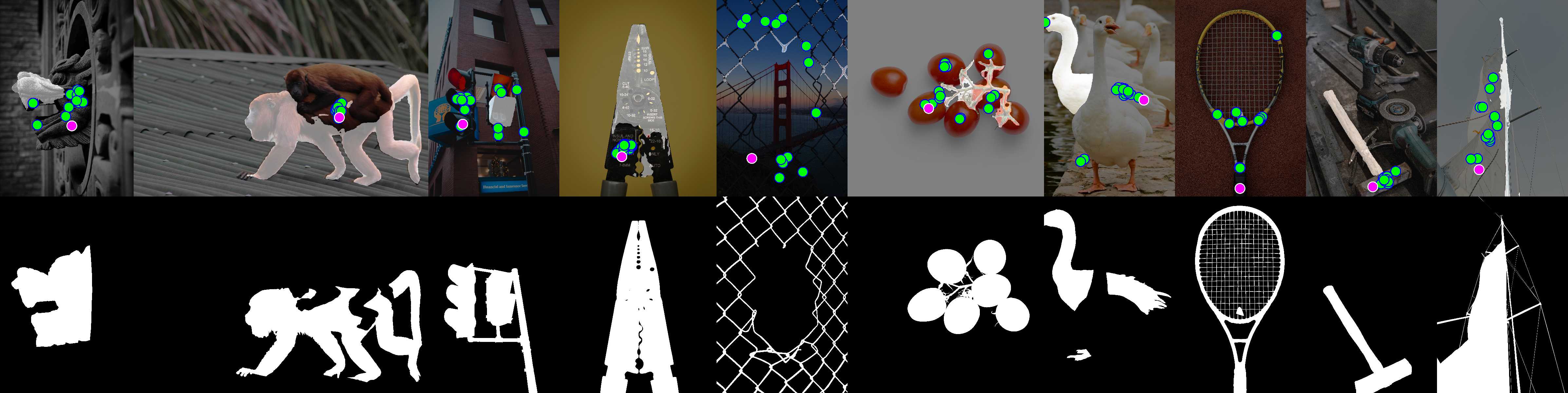}
    \includegraphics[width=1.0\textwidth]{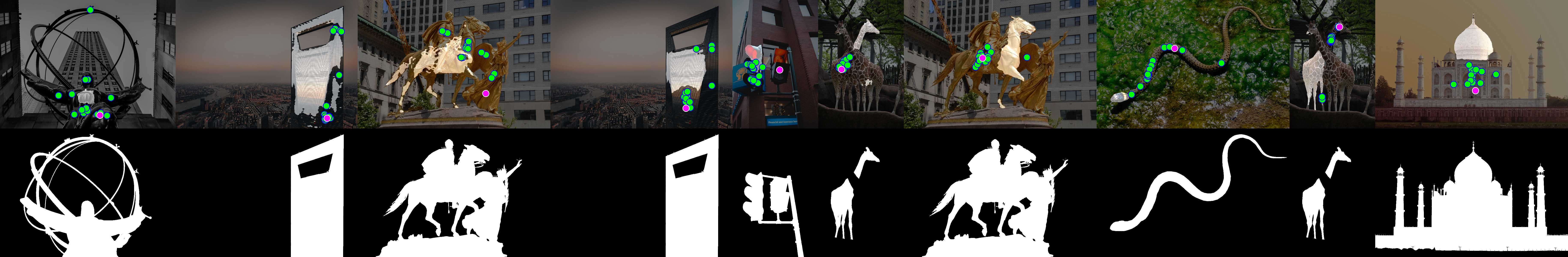}
    \includegraphics[width=1.0\textwidth]{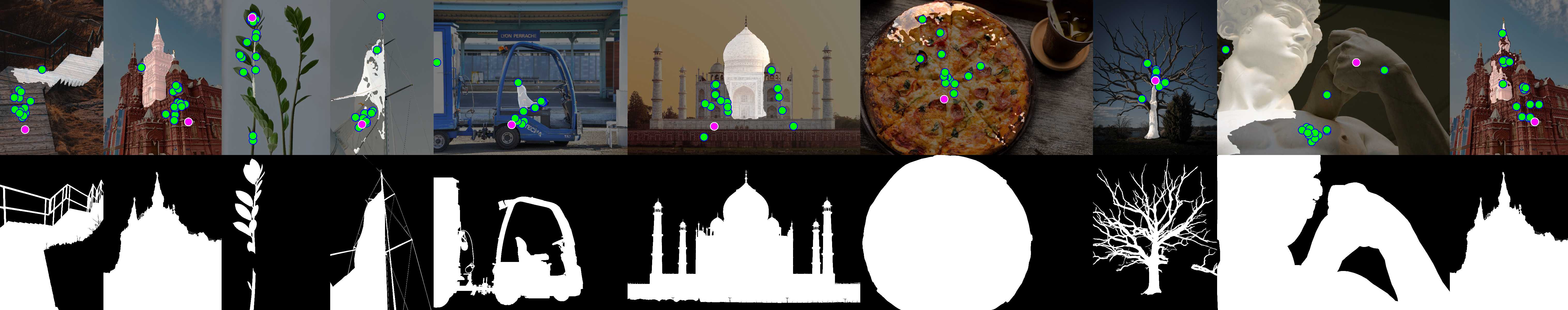}
    \caption{Top rows: images with white masks from previous interaction round (i.e. users asked to fix it) overlap, positive real-user clicks (green) and clicks, obtained from baseline strategy (magenta). Bottom rows: ground-truth masks.}
    \label{fig:second_positive}
\end{figure*}

\begin{figure*}
    \centering
    \includegraphics[width=1.0\textwidth]{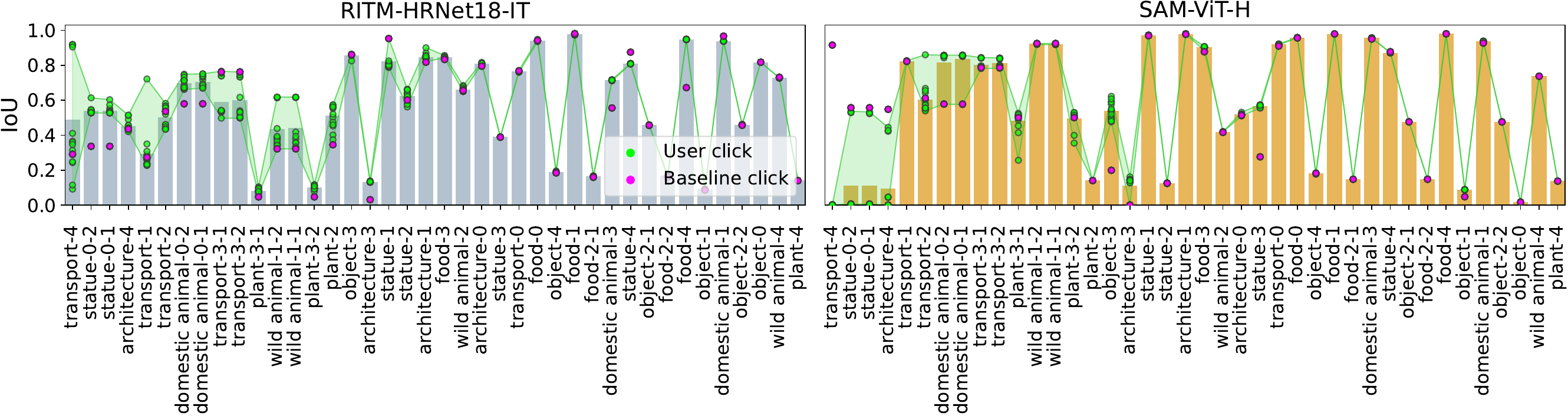}
    \caption{An IoU spread (a difference between a maximum and minimum IoU of user clicks) between predicted and ground truth masks in the second negative real user interaction round. Green points represent user clicks, magenta points depict the clicks generated with the baseline strategy. Columns are sorted by an average spread.}
    \label{fig:user-study-std31}
\end{figure*}

\begin{figure*}
    \centering
    \includegraphics[width=1.0\textwidth]{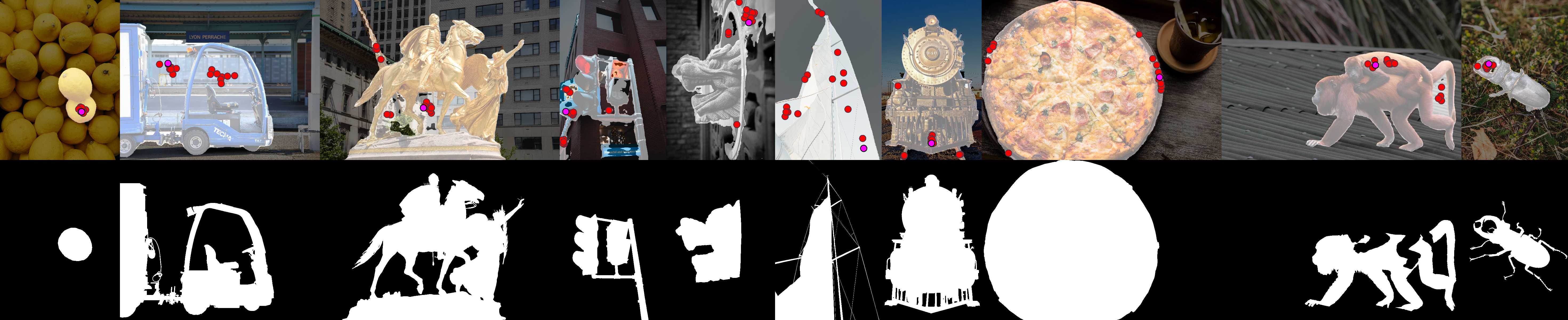}
    \includegraphics[width=1.0\textwidth]{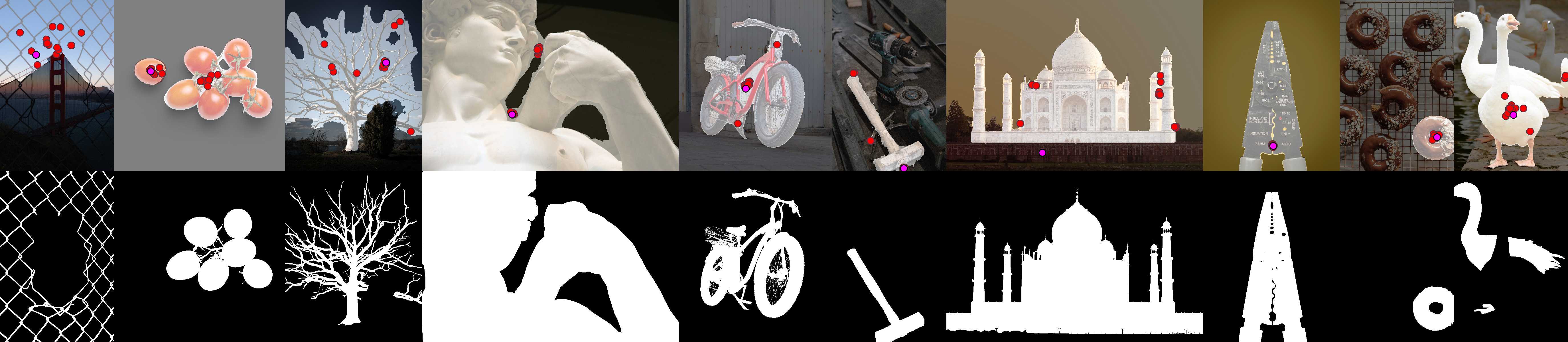}
    \includegraphics[width=1.0\textwidth]{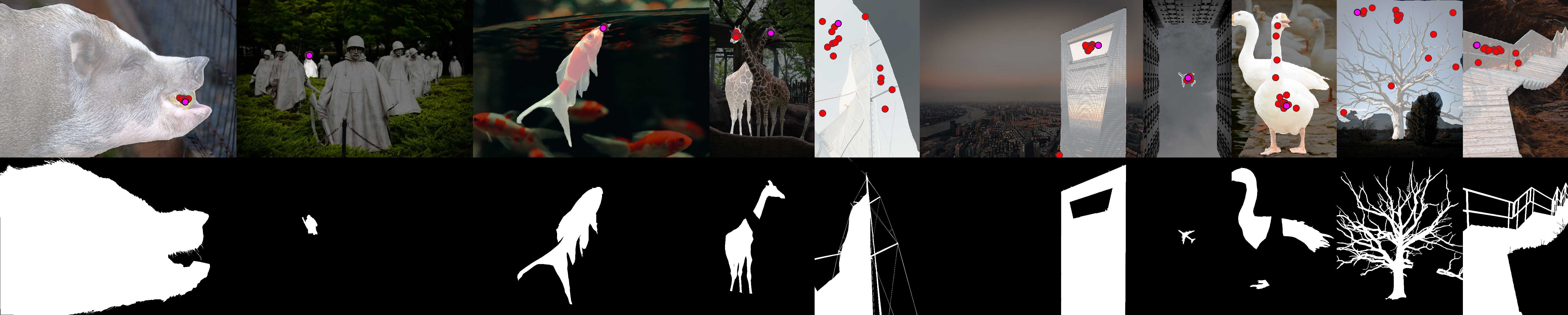}
    \includegraphics[width=1.0\textwidth]{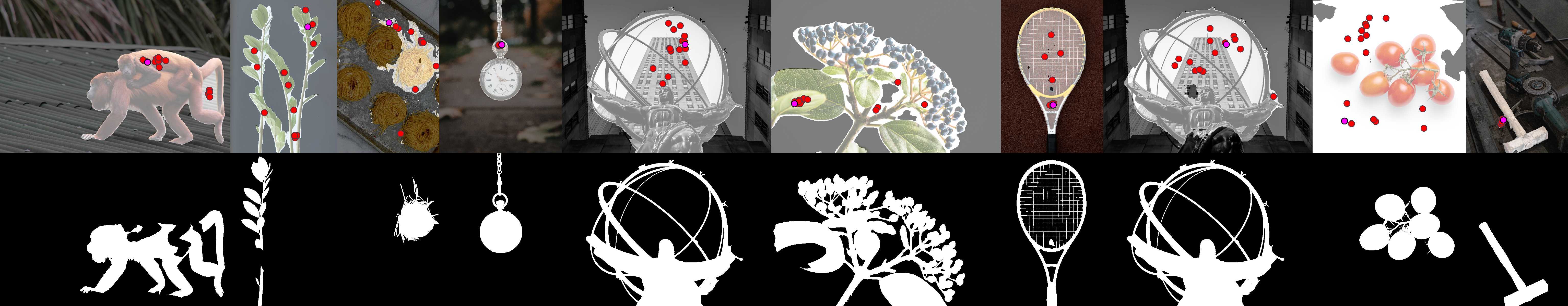}
    \caption{Top rows: images with white masks from previous interaction round (i.e. users asked to fix it) overlap, negative real-user clicks (red) and clicks, obtained from baseline strategy (magenta). Bottom rows: ground-truth masks.}
    \label{fig:second_negative}
\end{figure*}

\begin{figure*}
    \centering
    \includegraphics[width=1.0\textwidth]{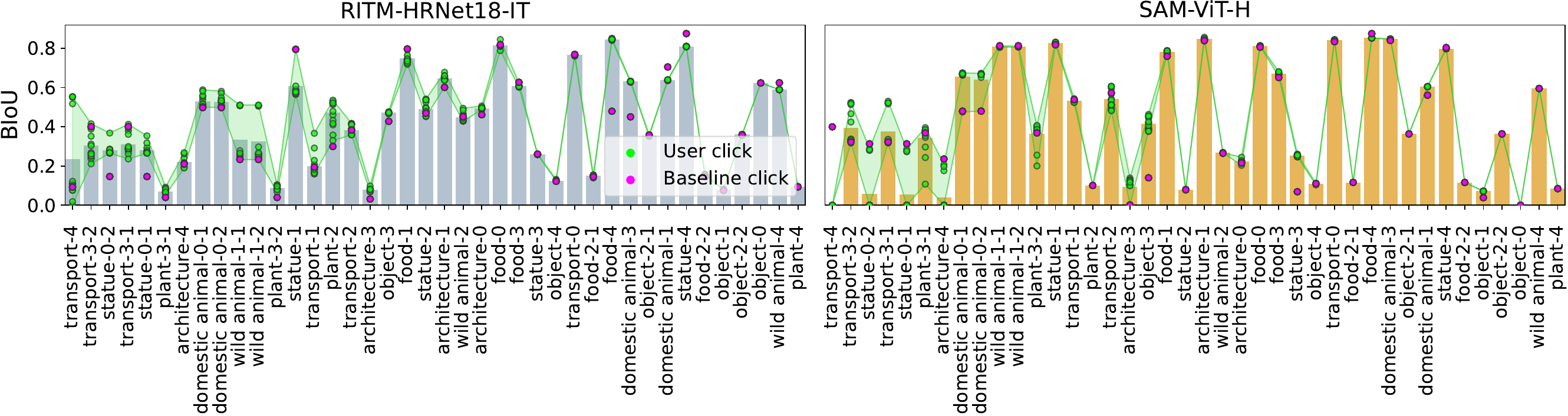}
    \caption{A BIoU spread (a difference between a maximum and minimum BIoU of user clicks) between predicted and ground truth masks in the second negative real user interaction round. Green points represent user clicks, magenta points depict the clicks generated with the baseline strategy. Columns are sorted by an average spread.}
    \label{fig:user-study-std32}
\end{figure*}

\begin{figure*}
    \centering
    \includegraphics[width=0.9\textwidth]{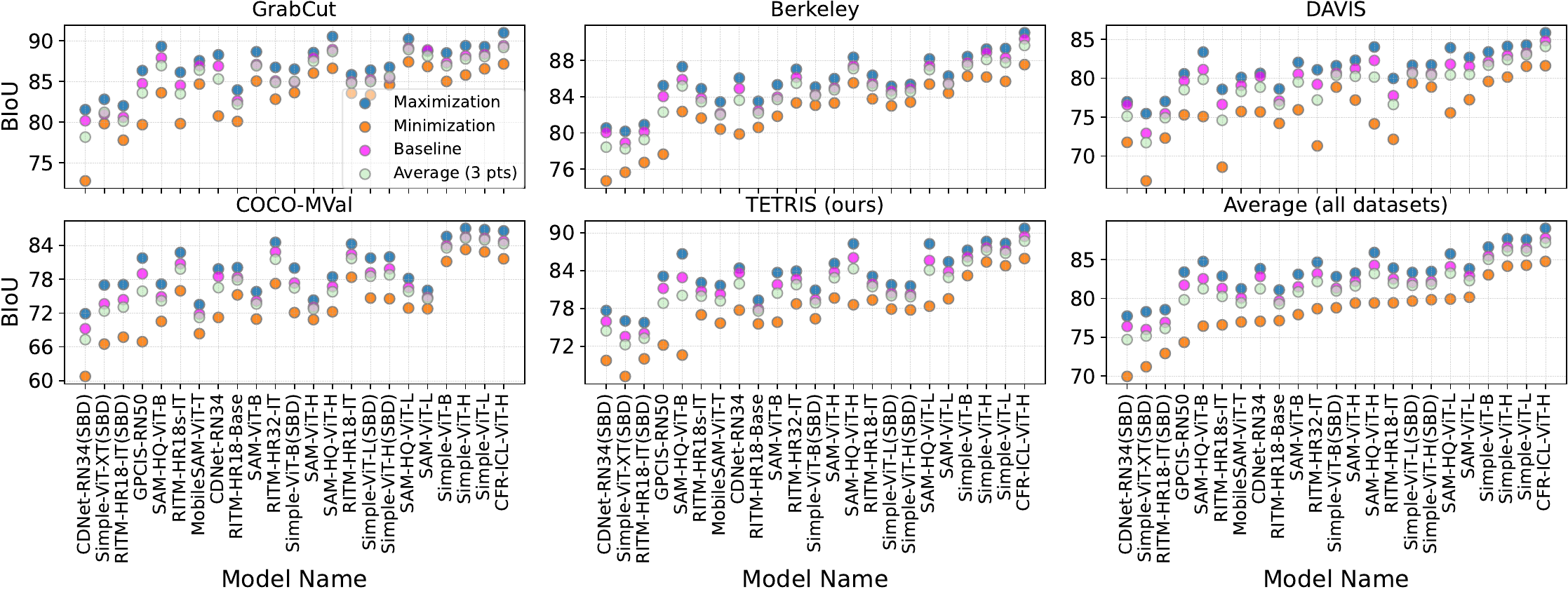}
    \caption{Visualization of BIoU bounds of tested models. Columns are sorted by an average BIoU-Min. As can be seen, there exists a dependency between the robustness (BIoU-D, a delta between BIoU-Max and BIoU-Min) and the prediction quality.} 
    \label{fig:biou_bounds}
    
    \centering
    \includegraphics[width=0.98\textwidth]{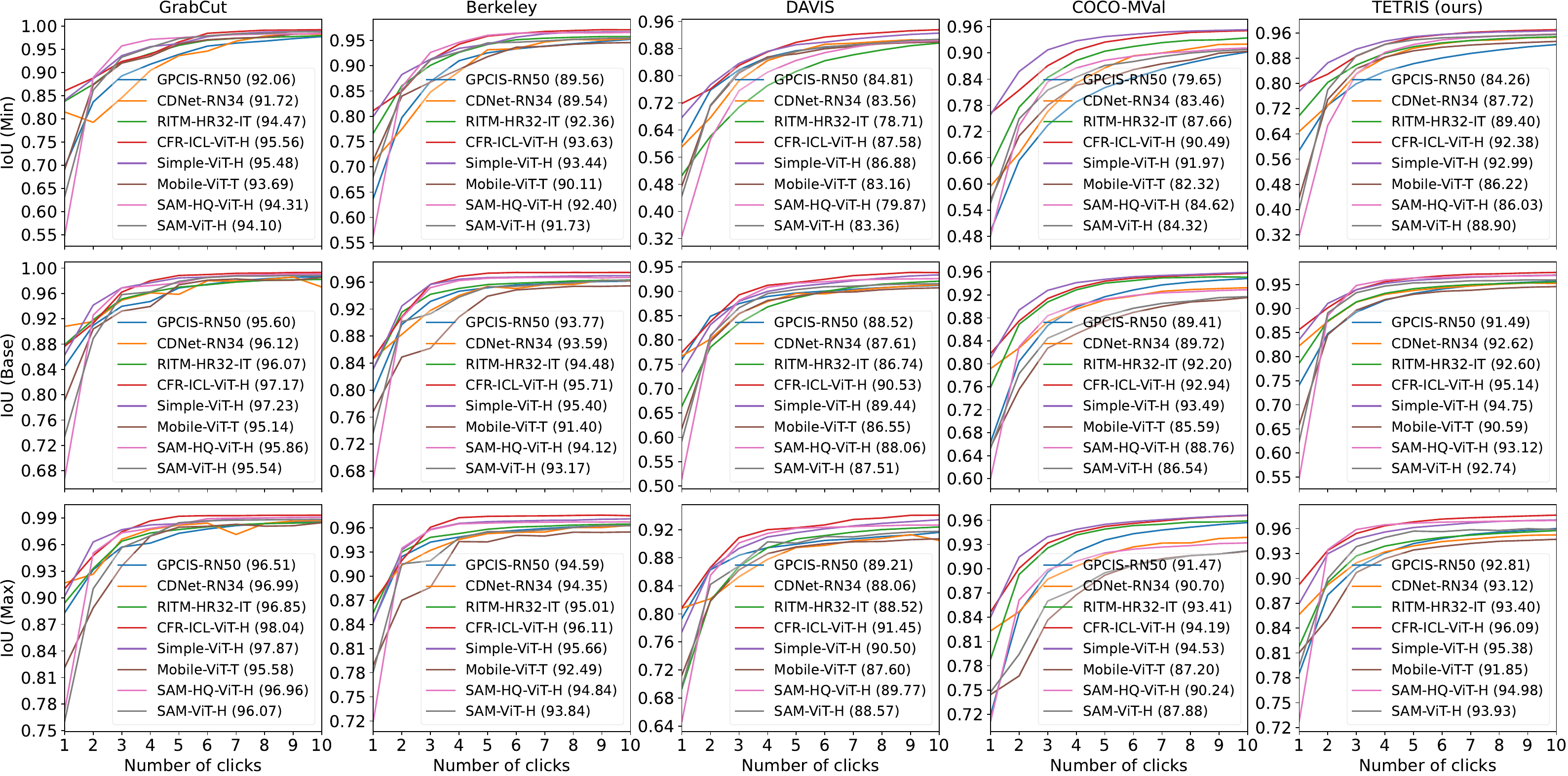}
    \includegraphics[width=0.98\textwidth]{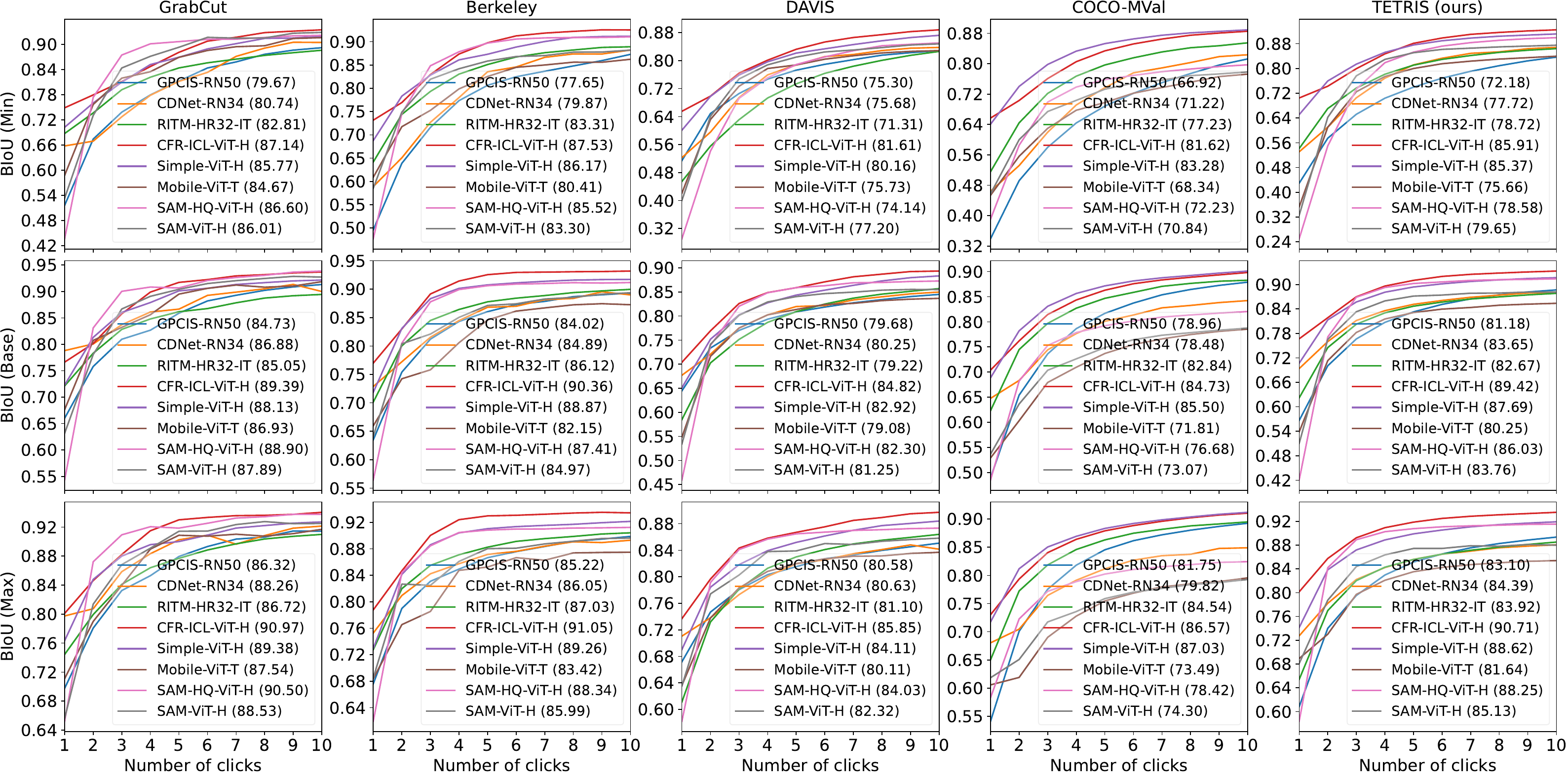}
    \caption{\textit{Minimizing, baseline} and \textit{maximizing trajectory} for IoU/BIoU metric. Aggregated value (AuC) presented in brackets.}
    \label{fig:plot1}
\end{figure*}

Figure~\ref{fig:matting-several-objects} presents an example of matting being successfully applied to an image with several objects. Expectedly, the uncertain regions are located around the boundaries between similar objects both having complex surfaces; after matting, these objects can be distinguished fairly well.

\section{Limitations of work}
\label{sec:bench-info}

We observe the following major limitations of our interactive segmentation robustness benchmark:
\begin{enumerate}
    \item \label{lim:1} During the evaluation, we optimize each adversarial user input for a fixed number of iterations, which does not guarantee global extreme finding and limits the scope.
    \item \label{lim:2} Using a greedy search of placing optimized clicks one by one, does not guarantee that the obtained trajectories will comply with common sense (e.g. \textit{maximization} trajectory should be above \textit{baseline}, and \textit{baseline} trajectory should be above \textit{minimization} trajectory on relevant charts).
    \item \label{lim:3} The proposed approach is much faster than an exhaustive search, however it is still an order of magnitude more computationally expensive than evaluation without optimization, since we utilize 10 forward-backward passes per click.
    \item \label{lim:4} Images in TETRIS-\textsc{things} subset are collected from the Unsplash photo website. Accordingly, using a single source of “raw” data might cause biases imposed by either technical parameters (specific camera models, high image quality) or image content (geographical location, object classes, object sizes, and locations of objects in an image). Both images and object classes are hand-picked (motivation and discussion of the selection strategy are expounded in Section~\ref{three_part}), which might be another source of biases in the dataset.
    \item \label{lim:5} The segmentation masks are manually labeled. The resulting annotations, even being extremely accurate, reflect a subjective opinion of an annotator and/or an expert annotator who verifies the results.
\end{enumerate}

To overcome Limitation~\ref{lim:1} we have proposed a way to explore all possible positions of user inputs using exhaustive search in pixel-wise grid, however it takes several hours per image and is not applicable to large datasets like TETRIS. Moreover, our empirical study shows that even small perturbation of click coordinates and local optimization greatly impacts models performance.

Abandoning greedy search from Limitation~\ref{lim:2} would require an exponential optimization runs w.r.t number of clicks, making comparison computationally impossible. Although the cases are that some trajectories on individual images are ordered inconsistently, in our study on all datasets aggregations we observe consistent ranking.

To overcome Limitation~\ref{lim:3} we utilize per-image parallelization since optimization of different images is performed independently.

We believe that Limitation~\ref{lim:4} is inevitable for a small dataset created through high-quality and resource-intensive data acquisition and annotation procedures. Moreover, TETRIS-\textsc{people} part was collected from various cameras, where the crowdsourcing performers captured themselves.

Limitation~\ref{lim:5} cannot be avoided when the “real” ground truth data is absent and targets need to be prepared manually. Usage of matting also seems unavoidable, since precise per-pixel labeling of objects with a complex surface (e.g., animals covered with fur) is extremely time-consuming.

\clearpage
\newpage
\bibliography{main}